\documentclass[preprint,12pt]{elsarticle}

\DeclareMathAlphabet{\mathscr}{U}{rsfs}{m}{n}
\usepackage{graphicx}%
\usepackage{multirow}%
\usepackage{amsmath,amssymb,amsfonts}%
\usepackage{amsthm}%
\usepackage{mathrsfs} 
\DeclareMathAlphabet{\mathscr}{U}{rsfs}{m}{n}
 \usepackage[title]{appendix}%
\usepackage{xcolor}%
\usepackage{textcomp}%
\usepackage{manyfoot}%
\usepackage{booktabs}%
\usepackage{algorithm}
\usepackage{algpseudocode}
\usepackage{listings}%
\usepackage{lmodern}
\usepackage{newtxtext, newtxmath}
\usepackage[hidelinks]{hyperref}
\usepackage{tabularx}
\usepackage{color}                        
\usepackage{caption}  
\usepackage{subcaption}
\usepackage[normalem]{ulem}  
\usepackage{xspace}   
\usepackage{natbib}

\usepackage{subcaption} 

\journal{Journal of Machine Learning for Computational Science and Engineering}

\begin{document}

\title{TabKAN: Advancing Tabular Data Analysis using Kolmogorov-Arnold Network} 

\author[1]{Ali Eslamian} \ead{ali.eslamian@uky.edu}
\author[2]{Alireza Afzal Aghaei} \ead{alirezaafzalaghaei@gmail.com}
\author[1,3]{Qiang Cheng \corref{cor1}} \ead{qiang.cheng@uky.edu}

\cortext[cor1]{Corresponding author: qiang.cheng@uky.edu}

\affiliation[1]{organization={Department of Computer Science, University of Kentucky},
            addressline={329 Rose Street}, 
            city={Lexington},
            postcode={40506}, 
            state={Kentucky},
            country={USA}}

\affiliation[2]{organization={Independent Researcher},
            city={Isfahan},
            country={Iran}}
\affiliation[3]{organization={Institute for Biomedical Informatics, University of Kentucky},
            addressline={800 Rose Street}, 
            city={Lexington},
            postcode={40506}, 
            state={Kentucky},
            country={USA}}


\begin{abstract}
Tabular data analysis presents unique challenges that arise from heterogeneous feature types, missing values, and complex feature interactions. While traditional machine learning methods like gradient boosting often outperform deep learning, recent advancements in neural architectures offer promising alternatives. In this study, we introduce TabKAN, a novel framework for tabular data modeling based on Kolmogorov–Arnold Networks (KANs). Unlike conventional deep learning models, KANs use learnable activation functions on edges, which improves both interpretability and training efficiency. TabKAN incorporates modular KAN-based architectures designed for tabular analysis and proposes a transfer learning framework for knowledge transfer across domains. Furthermore, we develop a model-specific interpretability approach that reduces reliance on post hoc explanations. Extensive experiments on public datasets show that TabKAN achieves superior performance in supervised learning and significantly outperforms classical and Transformer-based models in binary and multi-class classification. The results demonstrate the potential of KAN-based architectures to bridge the gap between traditional machine learning and deep learning for structured data.

\textbf{Code available on:} \href{https://github.com/aseslamian/TAbKAN}{https://github.com/aseslamian/TAbKAN}
\end{abstract}




\maketitle

\section{Introduction}
Tabular data, a fundamental form of structured information across domains such as healthcare, finance, and e-commerce, plays a central role in data-driven decision-making. Machine learning on tabular data has become increasingly important for scientific and engineering applications such as multiscale modeling and structural behavior prediction \cite{liu2025explainable, liu2021stochastic, liu2022stochasticB, liu2024stochastic, majidi2025predicting}. However, tabular data presents unique challenges such as heterogeneous feature types, missing values, non-stationary distributions, and complex inter-feature dependencies that make it difficult to design universally effective models.

Traditional machine learning methods, particularly tree-based ensembles such as gradient boosted, often outperform deep learning models on tabular datasets. Nonetheless, adapting deep architectures for tabular learning remains an active and important research area. Multi-Layer Perceptrons (MLPs) have been explored but are constrained by their use of fixed activation functions and limited capacity for modeling nonlinear feature interactions. Transformers, though powerful for sequential and textual data, often struggle to capture the structural and statistical heterogeneity of tabular data and typically offer limited interpretability.

Kolmogorov–Arnold Networks (KANs) have recently emerged as a promising alternative. Inspired by the Kolmogorov–Arnold representation theorem, KANs express any multivariate continuous function as a composition of univariate functions and summation operators. Unlike MLPs, which assign fixed nonlinearities to neurons, KANs place learnable activation functions on the edges, enabling flexible and data-adaptive modeling of feature relationships. This architectural design not only improves parameter efficiency and training robustness but also provides intrinsic interpretability, allowing visualization of how each feature contributes to the model output. These characteristics make KANs a natural and theoretically grounded fit for tabular data analysis.

This paper introduces TabKAN, a novel framework for modeling numerical and categorical features through KAN-based modules developed specifically for tabular data analysis. TabKAN incorporates various KAN-based architectures, including spline-KAN \cite{KAN}, ChebyKAN \cite{ss2024chebyshev}, Rational KAN (RKAN) \cite{aghaei2024rkan}, Fourier-KAN \cite{dong2024fan}, fractional-KAN (fKAN) \cite{fKAN}, and Fast-KAN \cite{FastKAN, ta2024bsrbf}, to flexibly adapt to diverse data characteristics and capture intricate statistical patterns. The diversity and heterogeneity of tabular datasets motivate the use of multiple KAN architectures, each offering distinct advantages in expressiveness, smoothness, and computational efficiency.

The primary contributions of this study are summarized as follows:
\begin{itemize}
    \item We introduce a family of modular KAN-based architectures tailored for tabular data analysis, enabling efficient modeling of both numerical and categorical features.
    \item We develop a transfer learning framework for KANs that facilitates effective knowledge transfer across heterogeneous domains.
    \item We propose model-intrinsic interpretability methods for tabular data learning, reducing reliance on post hoc explanation techniques.
    \item We provide a comprehensive empirical evaluation of supervised learning across binary and multi-class classification tasks on diverse benchmark datasets.
\end{itemize}

Experimental results demonstrate that TabKAN achieves stable and significantly improved performance in both supervised and transfer learning settings, consistently outperforming baseline models on multiple public datasets. By integrating the principles of the Kolmogorov–Arnold representation with modern neural design, TabKAN bridges the gap between traditional machine learning and deep learning, offering a robust, interpretable, and efficient solution for tabular data modeling.

\section{Related Work}
Existing methods for tabular learning face multiple obstacles, such as mismatched feature sets between training and testing, limited or missing labels, and the potential emergence of new features over time~\cite{maqbool2024model}. These methods can be categorized as:

\noindent\textbf{Classic Machine Learning Models.} Early techniques rely on parametric or non-parametric strategies like K-Nearest Neighbors (KNN), Gradient Boosting, Decision Trees, and Logistic Regression~\cite{Moderndeeplearning}. Popular models include Logistic Regression (LR), XGBoost~\cite{chen2016xgboost, zhang2020customer}, and MLP. A notable extension is the self-normalizing neural network (SNN)~\cite{klambauer2017self}, which uses scaled exponential linear units (SELU) to maintain neuron activations at zero mean and unit variance. While SNNs are simple and effective, they can fail on complex, high-dimensional data, which has led to the proposal of more advanced neural architectures.

\noindent\textbf{Deep Learning-Based Supervised Models.}
Building on Transformer architectures, methods such as AutoInt~\cite{song2019autoint} apply self-attention to learn feature importance, while TransTab~\cite{transtab} extends Transformers to handle partially overlapping columns across multiple tables. Such extensions support tasks like transfer learning, incremental feature learning, and zero-shot inference. TabTransformer~\cite{tabtransformer} applies self-attention to improve feature embeddings and achieves strong performance even with missing data. SAINT~\cite{SAINT} introduces hybrid attention at both row and column levels, pairs it with inter-sample attention and contrastive pre-training, and outperforms gradient boosting models including XGBoost~\cite{chen2016xgboost}, CatBoost~\cite{catboost}, and LightGBM~\cite{lightgbm} on several benchmarks. 

While these Transformer-based architectures have shown promise, their self-attention mechanisms were originally designed for sequential data and can be less transparent when modeling the specific, often non-linear interactions between heterogeneous tabular features. Similarly, MLPs, while effective, are limited by their reliance on fixed activation functions, which can lead to less parameter-efficient models for complex functions. The KAN-based framework we propose in this paper addresses these limitations directly. With learnable activation functions on network edges, KANs offer a more architecturally flexible and parameter-efficient alternative to MLPs. Furthermore, their foundation in the Kolmogorov-Arnold representation theorem provides a more direct and interpretable method for modeling feature relationships than the adapted attention mechanisms of Transformers. 

Other innovations include TabRet~\cite{tabret}, which implements a retokenization step for previously unseen columns, and XTab~\cite{xtab}, which provides for cross-table pretraining in a federated learning setup and handles heterogeneous column types and numbers. TabCBM~\cite{tabcbm} introduces concept-based explanations that support human oversight and balance predictive accuracy and interpretability. TabPFN~\cite{tabPFN} is a pretrained Transformer that performs zero-shot classification on tabular data through meta-learning, without requiring task-specific training. TabMap~\cite{yan2024interpretable} transforms tabular data into 2D topographic maps that encode feature relationships spatially and preserve values as pixel intensities. Such a structure helps convolutional networks detect association patterns efficiently and outperforms other deep learning-based supervised models. TabSAL~\cite{li2024tabsal} employs lightweight language models to generate privacy-free synthetic tabular data when raw data cannot be shared due to privacy concerns. TabMixer~\cite{eslamian2025tabmixer} builds on the MLP-mixer framework and captures both sample-wise and feature-wise interactions through a self-attention mechanism. In \cite{poeta2024benchmarking}, KAN-based models for tabular data were compared with MLPs, but the analysis was restricted to a baseline KAN architecture with a limited number of layers.

\section{Background: Kolmogorov-Arnold Networks (KANs)}
In this section, we first provide an overview of KANs, followed by a description of specific KAN-based architectures.

\subsection{Spline Kolmogorov-Arnold Network}
A general Kolmogorov-Arnold network (KAN) is defined as a composition of \(L\) Kolmogorov-Arnold layers. Given an input \(\mathbf{x}_0 \in \mathbb{R}^{n_0}\), the output is given by
\begin{equation}
\text{KAN}(\mathbf{x}_0)
= 
\bigl(\Phi_{L-1} \circ \cdots \circ \Phi_0\bigr)\,\mathbf{x}_0,
\end{equation}
where each \(\Phi_\ell\) denotes the \(\ell\)-th KAN layer and $\circ$ denotes a composition. The shape of the network is specified by an integer array \([n_0, n_1, \dots, n_L]\), with \(n_\ell\) representing the number of nodes in the \(\ell\)-th layer. The original Kolmogorov-Arnold representation \cite{liu2024kan} corresponds to a 2-layer KAN of shape \([n, 2n+1, 1]\). For a general case, denote the activation of the \(i\)-th node in layer \(\ell\) by \(x_{\ell,i}\). Between layers \(\ell\) and \(\ell+1\), there are \(n_\ell \times n_{\ell+1}\) univariate functions \(\phi_{\ell,j,i}\), each mapping an input from neuron \((\ell,i)\) to an intermediate output $\tilde{x}_{\ell,j,i} = \phi_{\ell,j,i}\bigl(x_{\ell,i}\bigr)$. The activation of neuron \((\ell+1,j)\) is then obtained by summing the contributions:
\begin{equation}
x_{\ell+1,j} 
= \sum_{i=1}^{n_\ell} \phi_{\ell,j,i}\bigl(x_{\ell,i}\bigr).
\end{equation}
In matrix notation, this becomes
\begin{equation}
\mathbf{x}_{\ell+1}
=
\begin{pmatrix}
\phi_{\ell,1,1}(\cdot) & \cdots & \phi_{\ell,1,n_\ell}(\cdot)\\
\vdots & \ddots & \vdots \\
\phi_{\ell,n_{\ell+1},1}(\cdot) & \cdots & \phi_{\ell,n_{\ell+1},n_\ell}(\cdot)
\end{pmatrix}
\mathbf{x}_\ell,
\end{equation}
where the matrix of functions \(\Phi_\ell\) defines the layer-wise transformation.

\subsection{Chebyshev Kolmogorov-Arnold Network (ChebyKAN)}

The ChebyKAN \cite{ss2024chebyshev} employs Chebyshev polynomials of the first kind, $\{T_k(x)\}_{k=0}^d$, to approximate nonlinear functions with fewer parameters than traditional MLPs. First, the input $\mathbf{x} \in \mathbb{R}^n$ is normalized to $[-1,1]$ with the hyperbolic tangent function:
\begin{equation}
\tilde{\mathbf{x}} = \tanh(\mathbf{x}).
\end{equation}
The Chebyshev polynomials are then computed up to degree $d$ using the recursive definition
\begin{align}
T_0(x) &= 1, \\
T_1(x) &= x, \\
T_k(x) &= 2x T_{k-1}(x) - T_{k-2}(x), \quad \text{for} \ k \geq 2.
\end{align}
This process creates a polynomial tensor $\mathbf{T}$. Let $\Theta \in \mathbb{R}^{n \times m \times (d+1)}$ be the trainable coefficient tensor for $n$ input features, $m$ outputs, and polynomial degree $d+1$. The output of the ChebyKAN layer is computed via Einstein summation:
\begin{equation} \label{eq:cheby}
   y_{bo} = \sum_{i=1}^{n} \sum_{k=0}^{d} T_{bik} \,\Theta_{iok} ,
\end{equation}
where $b$ indexes the batch. The optimization of $\Theta$ during training helps ChebyKAN learn a highly expressive mapping with exceptional accuracy and capitalizes on the orthogonality and rapid convergence of Chebyshev polynomials.  For the ChebyKAN architecture, we adopt a similar hyperparameter range to the KAN: the depth varies from \(1\) to \(10\); the number of neurons per layer ranges from \(5\) to \(100\) in increments of \(5\); and the polynomial order is chosen from the interval \([2,6]\).

\subsection{Fast Kolmogorov-Arnold Network (Fast KAN)}
FastKAN \cite{FastKAN} is a reengineered variant of KAN designed to significantly enhance computational efficiency by replacing the original 3\textsuperscript{rd}-order B-spline basis with Gaussian radial basis functions (RBFs). In this framework, Gaussian RBFs serve as the primary nonlinear transformation and effectively approximate the B-spline operations used in traditional KAN. In addition, it applies layer normalization \cite{ba2016layer} to keep inputs from drifting outside the effective range of these RBFs. Together, these adjustments simplify the overall design of FastKAN while preserving its accuracy. The output of an RBF network is a weighted linear combination of these radial basis functions. Mathematically, an RBF network with N centers can be expressed as:
\begin{equation}
f(x) \;=\; \sum_{i=1}^{N} w_{i}\,\phi\bigl(\|\mathbf{x} - \mathbf{c}_{i}\|\bigr),
\end{equation}
where \( w_i \) are the learnable parameters or coefficients, and \( \phi \) is the radial basis function, which depends on
the distance between the input $x$ and a center $c_i$ represented as:
\begin{equation}
\phi(r) \;=\; \exp\left(-\tfrac{r^2}{2 h^2}\right),
\end{equation}
While standard KAN consists of sums of univariate transformations to approximate multivariate functions, Fast KAN generalizes this principle in a deeper feedforward architecture. For an input vector $\mathbf{x} \in \mathbb{R}^d$, the output is computed as $\mathbf{y} \;=\; f_{L} \circ f_{L-1} \circ \cdots \circ f_{1}(\mathbf{x})$. For the FastKAN NAS, we set the depth between \(1\) and \(5\) and the number of neurons per layer between \(5\) and \(50\). These ranges were selected based on prior studies and preliminary experiments, balancing expressive capacity and computational efficiency to ensure robust model performance across varying levels of complexity. Our empirical search (Appendix A) consistently identified optimal configurations within these bounds, validating their appropriateness.


\subsection{Rational Kolmogorov-Arnold Network (rKAN)}

The Rational Kolmogorov–Arnold Network (RKAN) considers two rational-function extensions: the Padé Rational KAN (PadéRKAN), which is based on Padé approximation that represents functions as ratios of polynomials, and the Jacobi Polynomial KAN (JacobiKAN), which employs mapped Jacobi polynomials \cite{aghaei2024rkan}. 

\begin{equation}
R(x) = \frac{P_q(x)}{Q_k(x)} 
= \frac{\sum_{i=0}^q a_i\, x^i}{\sum_{j=0}^k b_j\, x^j}.
\end{equation}
In each PadéRKAN layer, this rational form acts as the activation function. Such a structure helps the model to capture asymptotic behavior and abrupt transitions with greater precision. Specifically, for an input $\mathbf{x} \in \mathbb{R}^d$, the layer outputs
\begin{equation}
\mathbf{y} = \frac{\sum_{i=0}^q \theta_i\, P_i(\mathbf{x})}{\sum_{j=0}^k \theta_j\, Q_j(\mathbf{x})},
\end{equation}
where $\theta_i$ and $\theta_j$ are learnable parameters for the numerator and denominator polynomials, respectively.

To optimize the architecture for rKAN, we select the following ranges for the PadéRKAN variant: the depth is chosen between \(1\) and \(5\); the number of neurons per layer ranges from \(5\) to \(100\) in steps of \(5\); the numerator order varies from \(2\) to \(6\); and the denominator order is also selected from the interval \([2,6]\).

\subsection{Fourier Kolmogorov-Arnold Network (Fourier KAN)}

Fourier KAN~\cite{xu2024fourierkan} uses a Fourier series expansion to capture both low- and high-frequency components in tabular or structured data. Given an input vector $\mathbf{x} \in \mathbb{R}^d$, the transformation function $\phi_F(\mathbf{x})$ introduces sine and cosine terms up to a grid size $g$, which gives the network a way to approximate highly complex or oscillatory functions. Formally,
\begin{equation}
\phi_F(\mathbf{x}) 
= \sum_{i=1}^d \sum_{k=1}^g 
\bigl( a_{ik} \cos(k \, x_i) 
+ b_{ik} \sin(k \, x_i) \bigr),
\end{equation}
where $a_{ik}$ and $b_{ik}$ are trainable coefficients. The hyperparameter $g$ controls the number of frequency components and balances representational power against computational cost.

A Fourier KAN layer applies this frequency-based feature mapping to each input dimension and then combines the resulting terms via learnable parameters. For example, an output neuron $y$ is computed as:
\begin{equation} \label{eq:fourier}
 y 
= \sum_{i=1}^d 
\sum_{k=1}^g
\Bigl( W_{ik}^{(c)} \cos(k \, x_i) 
+ W_{ik}^{(s)} \sin(k \, x_i) \Bigr)
+ b,
\end{equation}
where $W_{ik}^{(c)}$ and $W_{ik}^{(s)}$ are learnable weights for the cosine and sine terms, respectively, and $b$ is a bias. By using the orthogonality of trigonometric functions, Fourier KAN often achieves faster convergence than traditional MLPs and polynomial-based KANs while also reducing overfitting. For the FourierKAN architecture, we consider depths from \(1\) to \(5\), the number of neurons per layer ranging from \(5\) to \(50\), and grid sizes selected from the interval \([1,10]\).

\subsection{Fractional Kolmogorov-Arnold Network (fKAN)}

The Fractional Kolmogorov-Arnold Network (fKAN) \cite{aghaei2025fkan} incorporates fractional-order Jacobi functions into the Kolmogorov-Arnold framework to enhance expressiveness and adaptability. Each layer of fKAN uses a Fractional Jacobi Neural Block (fJNB), which introduces a trainable fractional parameter $\nu$ to adjust the polynomial basis dynamically. For an input $\mathbf{x}\in \mathbb{R}^d$, the fractional Jacobi polynomial $J_n^{(\alpha,\beta,\nu)}(x)$ is given by:
\begin{equation}
J_n^{(\alpha, \beta)}(x^\nu) = \frac{(\alpha + 1)_n}{n!} \sum_{k=0}^{n} \binom{n}{k} \frac{(\beta + 1)_{n-k}}{(\alpha + \beta + 1)_{n-k}} \left( \frac{x^\nu - 1}{2} \right)^k \left( \frac{x^\nu + 1}{2} \right)^{n-k},
\end{equation}
where $(\alpha,\beta)>-1$ determine the shape of the polynomial. Within fKAN, each layer applies a linear transformation followed by a fractional Jacobi activation. The structure helps the model to capture subtle data patterns. For the fKAN architecture, we set the depth between \(1\) and \(10\), the number of neurons per layer from \(5\) to \(100\) in steps of \(5\), and the polynomial order in the range \([2,6]\).

\subsection{Rational Kolmogorov-Arnold Network (RKAN)}

The Jacobi Rational Kolmogorov-Arnold Network (RKAN) \cite{rKAN} integrates Jacobi polynomials $J_n^{(\alpha,\beta)}(x)$ and a rational mapping $\phi(x, L) = \frac{x}{\sqrt{x^2+L^2}}$ to enhance nonlinear function approximation beyond the conventional $[-1,1]$ domain. For an input $\mathbf{x} \in \mathbb{R}^d$, the layer output is formulated as:
\begin{equation}
\mathbf{y} = \sum_{n=0}^N \theta_n\, J_n^{(\alpha,\beta)}(\phi(\mathbf{x}, L)),
\end{equation}
where $\theta_n$ and $L$ are trainable coefficients and $\alpha,\beta>-1$ specify the polynomial’s orthogonality weight function $\omega(x) = (1-x)^\alpha (1+x)^\beta$. The mapping $\phi(x, L)$ extends the polynomials to the infinite interval and makes data scaling needless. Similar to the fKAN, for architecture optimization, we set the depth between \(1\) and \(10\), the number of neurons per layer from \(5\) to \(100\) in steps of \(5\), and the polynomial order in the range \([2,6]\).
    
\section{Methodology}

\begin{figure}
    \centering
    \includegraphics[width=0.65\linewidth]{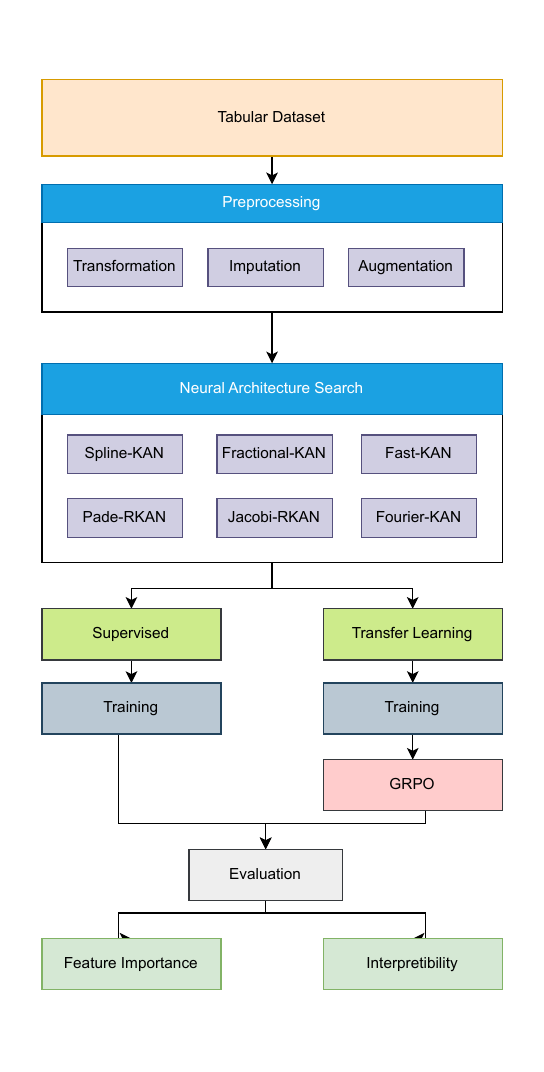}
    \caption{The structure of the TabKAN framework for tabular datasets.}
    \label{fig:tabkan}
\end{figure}

In this paper, we introduce TabKAN, a family of modular Kolmogorov–Arnold Network (KAN)-based architectures specifically engineered for tabular data. This family includes a diverse suite of models such as SplineKAN, ChebyKAN, JacobiRKAN, PadeRKAN, FourierKAN, fKAN, FastKAN, and their Mixer-enhanced variants. Our primary goals are to systematically optimize these models for both supervised and transfer learning tasks, employ Neural Architecture Search (NAS) to automatically identify optimal configurations, and use their functional formulation for inherent interpretability. The general schematic is shown in Figure~\ref{fig:tabkan}.

\subsection{Data Preprocessing}
To address missing values and class imbalance, we adopted the preprocessing strategy introduced in~\cite{eslamian2025tabmixer}. Let the input variable space be defined as $\mathcal{D}ata \in \{\mathbb{R} \cup \mathbb{C} \cup \mathbb{B} \cup \varnothing\}$, where $\mathbb{R}$, $\mathbb{C}$, and $\mathbb{B}$ denote the domains of numerical, categorical, and binary data, respectively. After the preprocessing block, we denote the resulting feature-target pair as $\{\mathcal{X}, \mathcal{Y} \}$, where $x$ contains numerical features, and $y$ is an integer used in classification tasks. The label set $\{\mathcal{Y}\}$ may have dimension one for binary classification or $M$ for multi-class classification.

Most tabular datasets contain both continuous numerical and categorical variables. We preprocess the categorical features by converting them into one-hot vectors. After preprocessing, the data is organized as an \( n \times m \) matrix with purely numerical entries (See Appendix \ref{prepocess_pipline} for more details).

\subsection{Neural Architecture Search}

Neural Architecture Search (NAS) aims to automatically identify optimal neural network configurations for a given learning task and replace manual design with a systematic search procedure. The effectiveness of NAS significantly depends on the strategy used to explore the candidate architecture space. Classical approaches such as grid or random search often suffer from combinatorial explosion or inefficient sampling. More advanced techniques, including Evolutionary Algorithms and Reinforcement Learning, can explore highly complex architecture spaces but are usually sample-inefficient and frequently require extensive training of numerous candidate models.

To mitigate this computational burden, we employ Bayesian Optimization (BO), which minimizes expensive evaluations of neural network performance by constructing a probabilistic surrogate model $f$ of the objective function. Typically instantiated as a Gaussian Process (GP), this surrogate model provides both a posterior mean $\mu(\mathbf{x})$ and a posterior standard deviation $\sigma(\mathbf{x})$ for any architecture $\mathbf{x}$. The choice of the next architecture for evaluation is guided by an acquisition function $\alpha(\mathbf{x})$, and balances exploitation (sampling near known optimal configurations) with exploration (sampling uncertain regions). A common acquisition function is \emph{Expected Improvement} (EI), defined as $\text{EI}(\mathbf{x}) = \mathbb{E}[\max(0, f(\mathbf{x}) - f(\mathbf{x}^+))]$, where $f(\mathbf{x}^+)$ represents the best performance observed thus far. The full algorithm is described in \ref{alg:gpbo}.

\begin{algorithm}[hb]
\caption{Gaussian Process-Based Bayesian Optimization}
\label{alg:gpbo}
\begin{algorithmic}[1]
\State \textbf{Input:} search space $\mathcal{X}$, objective function $f$, number of evaluations $N$
\State \textbf{Initialize:} sample $\{\mathbf{x}_i\}_{i=1}^{n_0}$ from $\mathcal{X}$; evaluate $y_i=f(\mathbf{x}_i)$
\For{$t = n_0+1$ \textbf{to} $N$}
    \State Fit GP on $\{(\mathbf{x}_i,y_i)\}_{i=1}^{t-1}$ to obtain $\mu(\mathbf{x}),\,\sigma(\mathbf{x})$
    \State Compute acquisition $\alpha(\mathbf{x})$ via EI:
    \Statex \(
      \mathrm{EI}(\mathbf{x})
      = \big(\mu(\mathbf{x}) - y^* - \xi\big)\,\Phi(Z)
        + \sigma(\mathbf{x})\,\phi(Z),\quad
      Z=\frac{\mu(\mathbf{x})-y^*-\xi}{\sigma(\mathbf{x})}
    \)
    \Statex where $y^*=\max_{1\le i < t} y_i$
    \State Solve $\mathbf{x}_t=\arg\max_{\mathbf{x}\in\mathcal{X}} \alpha(\mathbf{x})$
    \State Evaluate $y_t=f(\mathbf{x}_t)$
\EndFor
\State \textbf{Return} $\mathbf{x}_{\text{best}}=\arg\max_{1\le i \le N} y_i$
\end{algorithmic}
\end{algorithm}

In this study, we implement NAS using the Optuna framework~\cite{optuna}, which efficiently explores the search space through Bayesian optimization coupled with effective pruning strategies. For each KAN variant, we carry out a dedicated NAS procedure to determine the optimal combination of architecture and functional parameters:

For FastKAN, we tune the number of layers \(L\), the width vector \(\mathbf{w} = (w_1, \ldots, w_L)\), and the parameters of the RBF activation functions. In PadéRKAN, we optimize network depth, layer widths, and the polynomial degrees \((q, k)\). For FourierKAN, the grid size \(g\), which controls the frequency resolution of the Fourier expansion, is selected through NAS. The fKAN model includes hyperparameters such as depth, widths, and the Jacobi polynomial order. Finally, RKAN uses NAS to select depth, widths, and Jacobi polynomial order to adapt the rational architecture to varying dataset complexities.

We selected the Limited-memory Broyden–Fletcher–Goldfarb–Shanno (L-BFGS)  optimizer to guide the search. It is a quasi-Newton method that approximates the full Newton step, $\theta_{k+1} = \theta_k - \mathbf{H}_k^{-1} \nabla f(\theta_k)$, to guide the optimization process. All models are trained using the L-BFGS optimizer with cross-entropy loss. The BFGS algorithm iteratively builds an approximation $\mathbf{B}_{k+1}^{-1}$ to the inverse Hessian via the update rule:
\begin{equation}
    \mathbf{B}_{k+1}^{-1} = (\mathbf{I} - \rho_k s_k y_k^T) \mathbf{B}_k^{-1} (\mathbf{I} - \rho_k y_k s_k^T) + \rho_k s_k s_k^T, \quad \text{where } \rho_k = \frac{1}{y_k^T s_k}.
\end{equation}

L-BFGS avoids the $\mathcal{O}(n^2)$ memory cost of storing $\mathbf{B}_k^{-1}$ by using only the $m$ most recent update vectors: $s_k = \theta_{k+1} - \theta_k$ (the step) and $y_k = \nabla f(\theta_{k+1}) - \nabla f(\theta_k)$ (the change in gradient). These vectors implicitly define the quadratic model of the objective function. The search direction is computed efficiently via a two-loop recursion, which starts with an initial Hessian approximation, typically a scaled identity matrix $\mathbf{H}_k^0 = \gamma_k \mathbf{I}$, where the scaling factor is set as:
\begin{equation}
    \gamma_k = \frac{s_{k-1}^T y_{k-1}}{y_{k-1}^T y_{k-1}}.
\end{equation}
This formulation enables efficient second-order optimization while maintaining limited memory usage, making it well-suited for smooth, full-batch training landscapes such as those encountered in KAN models.

The validation F1 score served as the selection criterion for identifying optimal configurations, ensuring both generalization and adaptation to the structural and statistical characteristics of the data. To implement this, we performed a dedicated Neural Architecture Search (NAS) for each model-dataset pair using the Optuna framework. Each search consisted of 100 trials, where a proposed hyperparameter configuration was used to train a model and subsequently evaluated on the validation set. The configuration achieving the highest validation F1 score was selected as the optimal one. This final configuration was then retrained on the combined training and validation data and evaluated once on the held-out test set to report final performance. This systematic procedure ensured that every model was evaluated under its best-performing configuration, providing a fair and rigorous benchmark. Detailed results and analyses of the hyperparameter optimization procedures are presented in Appendix A.

\subsection{Supervised Learning}
In our supervised learning experiments, we evaluate various machine learning approaches categorized into classical baselines, specialized tabular models, and a suite of Kolmogorov-Arnold Network (KAN) variants. Classical baselines include Logistic Regression (LR), XGBoost, Multi-layer Perceptron (MLP), and Structured Neural Networks (SNN). Specialized tabular models evaluated include Attentive Interpretable Tabular Learning (TabNet), Deep Cross Network (DCN), Automatic Feature Interaction via Self-Attention (AutoInt), TabTransformer (TabTrans), Feature Tokenizer Transformer (FT-Trans), Variational Information Maximizing Exploration (VIME), Self-supervised contrastive learning using random feature corruption (SCARF), and Transferable Tabular
Transformers (TransTab). Additionally, we examine multiple KAN variants such as ChebyKAN, JacobiKAN, PadéRKAN, FourierKAN, fKAN, and fast-KAN, alongside the original KAN architecture.

Each model undergoes individual hyperparameter optimization tailored to its architectural characteristics and dataset-specific properties to ensure a fair and rigorous comparison. 

Models like wav-KAN \cite{wavKAN} and fc-KAN \cite{fc-kan}, although included in initial evaluations, demonstrated limitations. Wav-KAN consistently underperformed across datasets, while fc-KAN’s architectural complexity impeded practical deployment. For these reasons, both were ultimately excluded from our final comparative analysis.

\subsection{Transfer Learning}
With transfer learning, machine learning models can use knowledge learned from a source task to improve performance on a related target task through fine-tuning. While effective in domains with common structural patterns, such as computer vision and natural language processing, transfer learning for tabular data poses unique challenges. Issues such as feature heterogeneity, dataset-specific distributions, and a lack of universal structural characteristics often result in encoder overspecialization during conventional supervised pretraining. Models trained on classification objectives typically develop highly specialized representations suited to dominant patterns in the source dataset. Their adaptability to target tasks with varying feature spaces, class distributions, or differing objectives is therefore limited.

To systematically investigate these challenges, we adopt the methodological approach proposed by~\cite{transtab}. Specifically, we partition each dataset into two subsets, Set1 and Set2, with a controlled 50\% feature overlap. The setup simulates a cross-domain transfer learning scenario within each dataset, where overlapping features constitute shared knowledge, and non-overlapping features define distinct statistical domains. The controlled partial overlap provides a way to evaluate a model's ability to generalize existing representations while simultaneously adapting to new features.

The experimental procedure comprises two main stages: pretraining and fine-tuning. Initially, supervised training is performed on Set1 to establish robust initial feature representations. Upon reaching convergence, all layers except the final prediction layer (and any bias layers, if present) are frozen to preserve the learned patterns. In the subsequent fine-tuning phase, the unfrozen layers are trained with Set2, which makes the model adjust specifically to the target dataset's distribution. 

Additionally, we incorporate the Group Relative Policy Optimization (GRPO)~\cite{shao2024deepseekmath} method. It offers a robust fine-tuning mechanism for transfer learning and balances task-specific adaptation with knowledge retention. Its effectiveness is further analyzed in our ablation study. In certain scenarios, GRPO demonstrates improved performance over the standard fine-tuning procedure, which suggests its potential to further stabilize and refine feature transfer under domain shifts.

To thoroughly assess model robustness and bidirectional transfer, we perform evaluations on the test portion of Set2. Additionally, the roles of Set1 and Set2 are reversed in a cross-validation framework for a comprehensive examination of the model’s generalization capabilities under various domain shifts. The balanced approach helps overcome the inherent limitations posed by tabular data, such as feature heterogeneity and encoder overspecialization.

\subsection{KAN-Mixer Architecture}
To explore the integration of KAN into more advanced neural architectures, we adapted the MLP-Mixer framework. We replaced its standard MLP blocks with KAN layers, which resulted in the KAN-Mixer architecture~\cite{ibrahum2024resilient}. Such a modification retains the overall structure of TabMixer~\cite{eslamian2025tabmixer} and ensures compatibility with its attention and mixing components while using the representational power of KANs. The substitution of linear transformations with KAN-based approximators in the KAN-Mixer aims to enhance the expressivity and flexibility in modeling nonlinear patterns commonly observed in tabular datasets. The design choice provides for end-to-end differentiable training and incorporates the inductive biases introduced by the Kolmogorov-Arnold framework.

\section{Experiments and Results}

We evaluate our model on ten publicly available datasets across both supervised and transfer learning tasks. While multiple performance metrics are computed—AUC, F1 score, precision, and recall—we report only AUC due to its effectiveness in summarizing classification performance and limitations on space. To assess robustness, we compare our model with state-of-the-art baselines under varying data and feature configurations. Following the protocol in~\cite{transtab}, we use average ranking as the main comparison criterion, which provides an overall view of relative performance across datasets. All experiments run on an AMD Ryzen Threadripper PRO 5965WX 24-core CPU with 62 GB of RAM and an NVIDIA RTX A4500 GPU featuring 20 GB of memory.

\subsection{Datasets}

We employ a variety of datasets to evaluate our models, covering a broad spectrum of application areas:
\begin{itemize}
    \item \textbf{Financial Decision-Making:} Credit-g (CG) and Credit-Approval (CA) datasets
    \item \textbf{Retail:} Dresses-sale (DS) dataset, capturing detailed sales transactions
    \item \textbf{Demographic Analysis:} Adult (AD) and 1995-income (IO) datasets, containing income and census-related variables
    \item \textbf{Specialized Industries:}
    \begin{itemize}
        \item Cylinder bands (CB) dataset for manufacturing
        \item Blastchar (BL) dataset for materials science
        \item Insurance company (IC) dataset offering insights into the insurance domain
    \end{itemize}
\end{itemize}

Collectively, these benchmark datasets span diverse fields and data structures, which provides for a thorough assessment of our approach. Additional details for each dataset appear in Table~\ref{dataset}.

\begin{table*}[ht]
    \centering
    \caption{Dataset details including abbreviation, number of classes, number of data points, and number of features.}
    \label{dataset}
    \begin{tabular}{l c c c c}
        \toprule
        \textbf{Dataset Name} & \textbf{Abbreviation} & \textbf{\# Class} & \textbf{\# Data} & \textbf{\# Features} \\
        \midrule
        Credit-g                & CG  & 2  & 1,000 & 20 \\
        Credit-Approval         & CA  & 2  & 690 & 15 \\
        Dataset-Sales         & DS & 2  & 500 & 12 \\
        
        Adult         & AD  & 2  & 48,842 & 14 \\ \midrule
        Cylinder-Bands         & CB  & 2  & 540 & 35 \\
        Blastchar              & BL  & 2  & 7,043 & 35 \\
        Insurance-Co              & IO  & 2  & 5,822 & 85 \\
        1995-Income              & IC  & 2  & 32,561 & 14 \\ \midrule
        ImageSegmentation        & SG  & 7  & 2,310 & 20 \\
        ForestCovertype          & FO  & 7  & 581,012 & 55 \\ \bottomrule 
    \end{tabular}
\end{table*}


We choose the configuration that yields the highest validation performance and then train the model on each dataset using ten distinct random seeds to mitigate the impact of training variability. This procedure aligns with the comparative approach used in TabMixer \cite{eslamian2025tabmixer}.
To improve inference efficiency while preserving accuracy, we used PyTorch’s \texttt{torch.quantization} package to implement both static and dynamic post-training quantization, as well as quantization-aware training (QAT) \cite{kermani2025energy}. This reduced the memory footprint of some models by 3\% to 15\%, without a significant loss in accuracy.

\subsection{Baseline Models for Comparison}
We benchmark our proposed model against both classic and cutting-edge techniques, including \textbf{Logistic Regression (LR)}, \textbf{XGBoost} \cite{chen2016xgboost}, \textbf{MLP}, \textbf{SNN} \cite{klambauer2017self}, \textbf{TabNet} \cite{tabnet}, \textbf{DCN} \cite{wang2017deep}, \textbf{AutoInt} \cite{song2019autoint}, \textbf{TabTransformer} \cite{tabtransformer}, \textbf{FT-Transformer} \cite{fttrans}, \textbf{VIME} \cite{yoon2020vime}, \textbf{SCARF} \cite{bahri2021scarf}, \textbf{CatBoost} \cite{catboost}, \textbf{SAINT} \cite{SAINT}, and \textbf{TransTab} \cite{transtab}. These baselines span a range of approaches for tabular data, from traditional machine learning to the latest deep learning methods.

To ensure a fair comparison, we apply the same preprocessing and evaluation workflow across all models. After preprocessing, each dataset is divided into training, validation, and test sets with a 70/10/20 split. Crucially, all baseline models were subjected to the same rigorous hyperparameter optimization procedure described in Section 4.2. 

\subsection{Supervised Learning}
The experimental results, summarized in Table~\ref{table:supervised}, clearly illustrate performance distinctions among the evaluated models. ChebyKAN emerged as the highest-performing model across evaluated datasets. Its efficacy in capturing intricate decision boundaries underscores the stability and approximation properties of its Chebyshev polynomial basis.

KAN-based methods consistently outperformed conventional baseline models such as LR, XGBoost, MLP, and SNN, which highlights the advantages of adopting the Kolmogorov-Arnold framework. Furthermore, KAN variants frequently matched or exceeded performance levels of advanced transformer-based architectures (e.g., TabTrans, FT-Trans, and TransTab). The comparative advantage demonstrates the substantial expressive power of KAN models, particularly through specialized functional expansions.

The effectiveness of ChebyKAN, along with notable results from JacobiKAN, PadéRKAN, FourierKAN, fKAN, and fast-KAN, emphasizes the potential of polynomial, rational, and Fourier expansions to significantly enhance supervised learning tasks on tabular data. These findings reinforce the necessity of careful model selection and targeted hyperparameter tuning to maximize performance across diverse tabular datasets.

\begin{table*}[htbp]
\centering
\caption{Evaluation of Different Models for Supervised Learning} \label{table:supervised}
\resizebox{\textwidth}{!}{
\begin{tabular}{lcccccccc|c|c}
\hline
Methods & CG & CA & DS & AD & CB & BL & IO & IC & Rank (Std) $\downarrow$ & Average $\uparrow$ \\
\hline
Logistic Regression & 0.720 & 0.836 & 0.557 & 0.851 & 0.748 & 0.801 & 0.769 & 0.860 & 17 (2.45) & 0.768 \\
XGBoost & 0.726 & 0.895 & 0.587 & 0.912 & \textbf{0.892} & 0.821 & 0.758 & 0.925 & 9.06 (6.67) & 0.814 \\
MLP & 0.643 & 0.832 & 0.568 & 0.904 & 0.613 & 0.832 & 0.779 & 0.893 & 15.3 (3.13) & 0.758  \\
SNN & 0.641 & 0.880 & 0.540 & 0.902 & 0.621 & 0.834 & 0.794 & 0.892 & 13.6 (4.73) & 0.763 \\ \midrule
TabNet & 0.585 & 0.800 & 0.478 & 0.904 & 0.680 & 0.819 & 0.742 & 0.896 & 17.1 (3.49) & 0.738 \\
DCN & 0.739 & 0.870 & 0.674 & \underline{0.913} & 0.848 & 0.840 & 0.768 & 0.915 & 7.69 (4.12) & 0.821 \\
AutoInt & 0.744 & 0.866 & 0.672 & \underline{0.913} & 0.808 & \underline{0.844} & 0.762 & 0.916 & 7.94 (4.63) & 0.816 \\
TabTrans & 0.718 & 0.860 & 0.648 & \textbf{0.914} & 0.855 & 0.820 & 0.794 & 0.882 & 11.1 (5.85) & 0.811 \\ \midrule
FT-Trans & 0.739 & 0.859 & 0.657 & \underline{0.913} & \underline{0.862} & 0.841 & 0.793 & 0.915 &  8.19 (4.46) & 0.822 \\
VIME & 0.735 & 0.852 & 0.485 & 0.912 & 0.769 & 0.837 & 0.786 & 0.908 & 11.8 (4.58) & 0.786 \\
SCARF & 0.733 & 0.861 & 0.663 & 0.911 & 0.719 & 0.833 & 0.758 & 0.919 & 11 (4.56) & 0.800 \\
TransTab & 0.768 & 0.881 & 0.643 & 0.907 & 0.851 & 0.845 & 0.822 & 0.919 & \underline{6.88 (3.43)} & 0.830 \\ 
TabMixer & 0.660 & \textbf{0.907} & 0.659 & 0.900 & 0.829 & 0.821 & 0.974 & \textbf{0.969} &  7.94    (6.54) & 0.840 \\ \midrule
\textbf{KAN} & 0.806 & 0.870 & 0.616 & 0.907 & 0.739 & 0.844 & 0.956 & 0.902 & 8.69 (4.11) & 0.83  \\
\textbf{ChebyKAN} & 0.823 & 0.883 & 0.670 & 0.905 & \underline{0.862} & \textbf{0.859} & 0.951 & 0.905 & \textbf{5.88 (3.47)} & \textbf{0.857} \\
\textbf{JacobiRKAN} & \textbf{0.854} & 0.860 & 0.685 & 0.888 & 0.611 & 0.814 & \underline{0.957} & 0.885 & 11.5  (7.69) & 0.819 \\
\textbf{PadeRKAN} & 0.826 & 0.855 & 0.670 & 0.868 & 0.778 & 0.808 & 0.952 & 0.856 & 12.4 (6.52) & 0.827 \\
\textbf{Fourier KAN} & 0.771 & 0.870 & 0.650 & 0.906 & 0.820 & 0.649 & 0.879 & \underline{0.935} & 9.31  (5.08) & 0.810 \\
\textbf{fKAN} & \underline{0.848} & 0.870 & \textbf{0.691} & 0.892 & 0.692 & 0.811 & 0.954 & 0.890 & 10.2  (6.64) & 0.831 \\
\textbf{fast-KAN} & \textbf{0.854} & \underline{0.897} & \underline{0.688} & 0.892 & 0.767 & 0.837 & \textbf{0.960} & 0.887 & 7.44 (6.55) & \underline{0.848} \\

\midrule
\end{tabular}
}
\end{table*}

Supervised learning requires ample labeled data; however, recent studies improve analysis using hybrid domain-specific methods \cite{deldadehasl2025customer}, multimodal approaches that combine language models with tabular inputs \cite{su2024tablegpt2}, or integrations of vision and tabular data for medical prediction tasks \cite{huang2023multimodal}.

\subsection{Transfer Learning}
We evaluate various KAN-based architectures and baseline models with the described transfer learning methodology. The results, summarized in Table~\ref{table:transfer}, demonstrate clear performance advantages among specific KAN variants.

FourierKAN emerges as the highest-performing KAN architecture, with an average performance of 0.859, and ranks second overall among all evaluated models. The performance surpasses not only classical approaches such as XGBoost (0.776) and MLP (0.775) but also Transformer-based methods including TabTransformer (0.764), AutoInt (0.754), and DCN (0.758). FourierKAN’s superior adaptability is attributed to its Fourier series expansion, where smooth, periodic basis functions effectively approximate both low- and high-frequency components in data distributions and facilitate robust adaptation to shifting feature domains.

Other KAN variants, such as JacobiKAN (0.814), ChebyKAN (0.796), and the base KAN model (0.774), also yield strong performances and frequently exceed conventional baseline approaches. The consistently strong results across these variants underscore the effectiveness of KAN models in addressing the complexities associated with tabular transfer learning. Notably, JacobiKAN’s orthogonal polynomial basis and ChebyKAN’s minimax approximation properties significantly contribute to their robust performance. This fact indicates the value of diverse functional approximations within the KAN family in handling domain-specific variability.

\begin{table*}[htbp]

  \centering
  \caption{Evaluation of Models for Transfer Learning} \label{table:transfer}
  \resizebox{\textwidth}{!}{
  \begin{tabular}{lcccccccccccccccc|c|c}
    \toprule
    \multirow{2}{*}{Methods} & \multicolumn{2}{c}{CG} & \multicolumn{2}{c}{CA} & \multicolumn{2}{c}{DS} & \multicolumn{2}{c}{AD} & \multicolumn{2}{c}{CB} & \multicolumn{2}{c}{BL} & \multicolumn{2}{c}{IO} & \multicolumn{2}{c}{IC} & \multicolumn{1}{c}{Rank(Std)  $\downarrow$ } & \multicolumn{1}{c}{Average  $\uparrow$ } \\ 
    & set1 & set2 & set1 & set2 & set1 & set2 & set1 & set2 & set1 & set2 & set1 & set2 & set1 & set2 & set1 & set2 \\ \midrule
    Logistic Regression & 0.69 & 0.69 & 0.81 & 0.82 & 0.47 & 0.56 & 0.81 & 0.81 & 0.68 & 0.78 & 0.77 & 0.82 & 0.71 & 0.81 & 0.81 & 0.84 & 14.5 (2.82) & 0.736 \\
    XGBoost & 0.72 & 0.71 & 0.85 & 0.87 & 0.46 & 0.63 & 0.88 & \underline{0.89} & \underline{0.80} & 0.81 & 0.76 & 0.82 & 0.65 & 0.74 & 0.92 & \underline{0.91} & 9.53 (5.38) & 0.776  \\
    
    MLP & 0.67 & 0.70 & 0.82 & 0.86 & 0.53 & \underline{0.67} & \underline{0.89} & \textbf{0.90} & 0.73 & \underline{0.82} & 0.79 & 0.83 & 0.70 & 0.78 & 0.90 & 0.90 & 9.84 (4.23) & 0.775 \\ 
    SNN & 0.66 & 0.63 & 0.85 & 0.83 & 0.54 & 0.42 & 0.87 & 0.88 & 0.57 & 0.54 & 0.77 & 0.82 & 0.69 & 0.78 & 0.87 & 0.88 & 14.5 (3.90) & 0.727 \\ \midrule 
    TabNet & 0.60 & 0.47 & 0.66 & 0.68 & 0.54 & 0.53 & 0.87 & 0.88 & 0.58 & 0.62 & 0.75 & 0.83 & 0.62 & 0.71 & 0.88 & 0.89 & 15.9 (4.09) & 0.692  \\
    DCN & 0.69 & 0.70 & 0.83 & 0.85 & 0.51 & 0.58 & 0.88 & 0.74 & 0.79 & 0.78 & 0.79 & 0.76 & 0.70 & 0.71 & 0.91 & 0.90 & 11.4  (4.51) & 0.758  \\
    AutoInt & 0.70 & 0.70 & 0.82 & 0.86 & 0.49 & 0.55 & 0.88 & 0.74 & 0.77 & 0.79 & 0.79 & 0.76 & 0.71 & 0.72 & 0.91 & 0.90 & 11.6  (4.39) & 0.754 \\ 
    TabTrans & 0.72 & 0.72 & 0.84 & 0.86 & 0.54 & 0.57 & 0.88 & \textbf{0.90} & 0.73 & 0.79 & 0.78 & 0.81 & 0.67 & 0.71 & 0.88 & 0.88 & 11.5 (3.57) & 0.764 \\  \midrule
    FT-Trans & 0.72 & 0.71 & 0.83 & 0.85 & 0.53 & 0.64 & \underline{0.89} & 0.90 & 0.76 & 0.79 & 0.78 & 0.84 & 0.68 & 0.78 & 0.91 & \underline{0.91} & 8.84 (3.82) & 0.781 \\ 
    VIME & 0.59 & 0.70 & 0.79 & 0.76 & 0.45 & 0.53 & 0.88 & \textbf{0.90} & 0.65 & 0.81 & 0.58 & 0.83 & 0.67 & 0.70 & 0.90 & 0.90 & 14.5  (5.37) & 0.718 \\
    SCARF & 0.69 & 0.72 & 0.82 & 0.85 & 0.55 & 0.64 & 0.88 & \underline{0.89} & 0.77 & 0.73 & 0.78 & 0.83 & 0.71 & 0.75 & 0.90 & 0.89 & 10.1  (2.87) & 0.778 \\ 
    
    TransTab & 0.74 & 0.76 & 0.87 & \textbf{0.89} & 0.55 & 0.66 & 0.88 & \textbf{0.90} & \underline{0.80} & 0.80 & 0.79 & 0.84 & 0.73 & 0.82 & 0.91 & \underline{0.91} & 5.56  (2.17) & 0.803 \\ 

    TabMixer & \textbf{0.86} & \underline{0.84} & \underline{0.87} & \underline{0.88} & 0.64 & \textbf{0.71} & \textbf{0.90} & \textbf{0.90} & \textbf{0.94} & 0.77 & \textbf{0.93} & \textbf{0.92} & \textbf{0.95} & \textbf{0.95} & \underline{0.94} & \textbf{0.95} & \textbf{1.91 (1.14)} & \textbf{0.883} \\ \midrule

    \textbf{KAN} & 0.80 & 0.81 & 0.86 & 0.86 & 0.50 & 0.50 & 0.56 & 0.64 & 0.73 & 0.74 & 0.84 & 0.85 & \textbf{0.95} & \textbf{0.95} & 0.90 & 0.90 & 9.19 (6.18) & 0.774 \\
    
    \textbf{ChebyKAN} & 0.79 & 0.76 & \textbf{0.89} & \textbf{0.89} & 0.60 & 0.60 & 0.84 & 0.88 & 0.77 & 0.50 & 0.65 & \underline{0.86} & \underline{0.91} & \underline{0.89} & 0.82 & 0.82 & 8.38 (5.71) & 0.796 \\ 

    \textbf{JacobiKAN} & \underline{0.85} & \textbf{0.86} & 0.85 & 0.86 & \underline{0.66} & \underline{0.68} & 0.86 & 0.88 & 0.61 & 0.61 & 0.82 & 0.82 & \textbf{0.95} & \textbf{0.95} & 0.88 & 0.88 & 8.28  (5.88) & 0.814 \\
    
    \textbf{PadeRKAN} & 0.76 & 0.77 & 0.87 & 0.80 & 0.50 & 0.62 & 0.86 & 0.50 & 0.64 & 0.64 & 0.66 & 0.66 & 0.88 & 0.76 & 0.63 & 0.50 & 13.7  (5.51) & 0.691 \\

    \textbf{Fourier KAN} & 0.83 & 0.82 & \textbf{0.89} & \underline{0.88} & \textbf{0.67} & \underline{0.68} & \textbf{0.90} & \textbf{0.90} & \underline{0.86} & \textbf{0.86} & \underline{0.85} & 0.85 & \textbf{0.95} & \textbf{0.95} & \textbf{0.95} & 0.90 & \underline{2.72 (1.56)} & \underline{0.859} \\

    \textbf{fKAN} & 0.76 & 0.74 & 0.82 & 0.78 & 0.57 & 0.58 & 0.68 & 0.78 & 0.60 & 0.63 & 0.64 & 0.68 & 0.80 & 0.77 & 0.74 & 0.72 & 14.2 (4.77) & 0.704 \\
    
    \textbf{Fast-KAN} & 0.71 & 0.81 & 0.84 & 0.75 & 0.57 & 0.53 & 0.66 & 0.71 & 0.63 & 0.62 & 0.73 & 0.70 & \underline{0.89} & 0.85 & 0.70 & 0.70 & 13.8 (5.53) & 0.713 \\  \bottomrule

  \end{tabular} 
  }
\end{table*}

\subsection{Multi-class Classification}
Table~\ref{table:multi} presents a comparison between TabKAN and several neural network baselines on two multi-class classification benchmarks. Since these tasks often involve class imbalance, macro-F1 was selected as the primary evaluation metric during training to ensure balanced performance across all classes \cite{TabKANet}. All KAN variants consistently outperform baseline models, with JacobiKAN achieving the highest overall performance. Its use of Jacobi polynomials, parameterized by $\alpha$ and $\beta$, provides a more adaptable polynomial basis, which supports improved approximation of complex patterns. TabTrans does not have the capability to handle categorical input, so we could not run the SA dataset on it \cite{TabKANet}.

\begin{table*}[htbp]
\centering
\caption{Comparison of different methods on SG and FO datasets.}
\label{table:multi}
\begin{tabular}{l|cc|cc|c}
\toprule
\multirow{2}{*}{Methods} & \multicolumn{2}{c|}{SA} & \multicolumn{2}{c|}{FO} & \multirow{2}{*}{Rank $\downarrow$} \\
 & ACC & F1 & ACC & F1 &  \\
\midrule
MLP & 90.97 & 90.73 & 67.09 & 48.03 & 9.25 (0.5)  \\
TabTrans  & - & - & 68.76 & 49.47 & 8.5 (0.707)  \\
TabNet  & 96.09 & 94.96 & 65.09 & 52.52 & 7.25 (2.5)  \\ 
\midrule
\textbf{KAN} & 96.32 & 96.33 & 85.11 & 84.80 & 4  (1.15) \\
\textbf{ChebyKAN} & \textbf{96.54} & \textbf{96.54} & 82.67 & 82.38 & 4  (3.46) \\
\textbf{JacobiKAN} & \underline{96.49} & \underline{96.49} & \textbf{96.56} & \textbf{96.56} & \textbf{1.5 (0.577)} \\
\textbf{PadeRKAN} & 94.81 & 94.78 & 92.95 & 92.94 & 5.5  (2.89) \\
\textbf{Fourier KAN}  & 95.89 & 95.89 & 84.55 & 84.42 & 5.62 (0.479) \\
\textbf{fKAN} & 95.89 & 95.93 & \underline{95.80} & \underline{95.79} & \underline{3.38 (1.70)} \\
\textbf{fast-KAN} & 95.45 & 95.44 & 87.13 & 86.98 & 5.25 (1.5)  \\
\bottomrule
\end{tabular}
\end{table*}

\subsection{Interpretability}

Interpretability in machine learning has two general approaches: model-specific methods and model-agnostic methods. Model-specific techniques are tailored to a given architecture, such as the interpretation of coefficients in linear regression as indicators of feature importance. In contrast, model-agnostic methods (e.g., SHAP, LIME, PDP) can be applied to any model but typically operate as post hoc approximations, which may introduce additional assumptions and reduce reliability.  

A key strength of Kolmogorov–Arnold Networks (KANs) is their \textit{built-in interpretability}. Unlike traditional black-box models (e.g., deep neural networks or gradient-boosted trees), KANs represent each connection between a feature and a hidden unit as a univariate function parameterized by well-defined mathematical bases. These functions can be reconstructed after training and visualized directly for architecture-driven explanations without requiring external surrogate models. Each feature is thus transformed by a learnable function that is directly accessible after training. Such a design gives a way to visualize feature-wise contributions and functional mappings without resorting to external interpretability tools. 

In ChebyKAN, feature transformations are Chebyshev polynomial expansions,
\begin{equation}
    f_{\text{Cheb}}(x) \;=\; \sum_{k=0}^{d} c_k\, T_k(x)
\end{equation}

where $T_k$ are Chebyshev polynomials and $c_k$ are learned coefficients. After inputs are normalized to $[-1,1]$, the resulting function can be visualized directly to reveal feature contributions. Linear or monotone shapes correspond to proportional influences, whereas oscillatory curves indicate more complex nonlinear effects.

FourierKAN instead employs a truncated Fourier expansion,
\begin{equation}
f_{\text{Fourier}}(x) \;=\; \sum_{k=1}^{K} \big( a_k \cos(kx) + b_k \sin(kx) \big),
\end{equation}
with coefficients ${a_k, b_k}$ learned during training. The superposition of sinusoidal terms gives the model a way to encode periodic and oscillatory dependencies. Visualizing these expansions exposes whether a feature contributes through periodicities, thresholds, or smooth monotonic trends. The representation is especially interpretable in domains with cyclic structure.

PadéRKAN generalizes this framework and models feature transformations as rational functions,
\begin{equation}
f_{\text{Pade}}(x) \;=\; \frac{P(x)}{Q(x)}, \quad
P(x) = \sum_{i=0}^{m} w^{(P)}_i \, \Phi^{(P)}_i(x), \quad
Q(x) = \sum_{j=0}^{n} w^{(Q)}_j \, \Phi^{(Q)}_j(x),
\end{equation}
where $\Phi^{(P)}_i$ and $\Phi^{(Q)}_j$ are shifted Jacobi polynomial bases with learned coefficients. Inputs are mapped to $[0,1]$ via a sigmoid, and the reconstructed rational maps can be plotted post-training. The resulting visualizations reveal sharp transitions, asymptotic trends, and non-polynomial patterns not easily captured by additive bases. To avoid artifacts near zeros of $Q(x)$, a small denominator floor can be applied.

In our framework, each feature’s univariate function offers non-parametric insights into its role in prediction. Visualizations can reveal monotonic trends, thresholds, or saturation effects that align with known domain behavior. Moreover, while KANs model features through univariate functions, deeper layers combine these representations additively, which creates complex multivariate dependencies. Co-variations among learned functions of related features may reflect latent interactions and provide further avenues for domain-informed interpretation.  

Figures~\ref{fig:1}, \ref{fig:2}, and \ref{fig:3} illustrate the attributions of feature A2 in the CA dataset, while Figures~\ref{fig:4}, \ref{fig:5}, and \ref{fig:6} depict the attributions of feature B. Figures~\ref{fig:1} and \ref{fig:4} demonstrate the interpretability of FourierKAN, whereas Figures~\ref{fig:2} and \ref{fig:5} highlight ChebyKAN. The differences in scale relative to Partial Dependence Plots (PDPs) arise from input normalization. The plotted functions reveal not only monotonic relationships and threshold effects but also oscillatory patterns (in FourierKAN) and asymptotic behaviors (in PadeRKAN).

\begin{figure}[htbp]
    \centering
    \begin{subfigure}[b]{0.45\textwidth} 
    \scalebox{1.05}{
        \includegraphics[width=\linewidth]{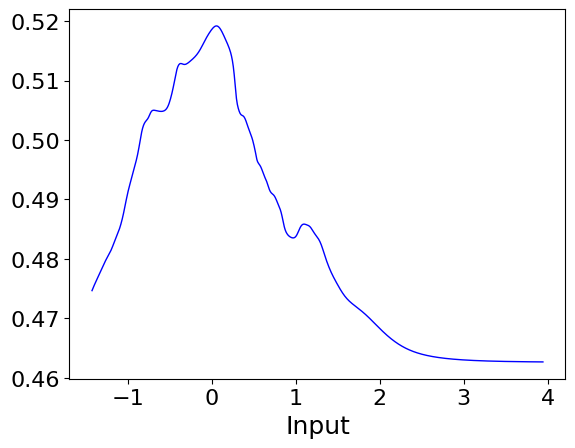}
        }
        \caption{Partial Dependence Plot - feature A} \label{fig:1}
    \end{subfigure}
    \hfill
    \begin{subfigure}[b]{0.45\textwidth}
    \scalebox{1.05}{
        \includegraphics[width=\linewidth]{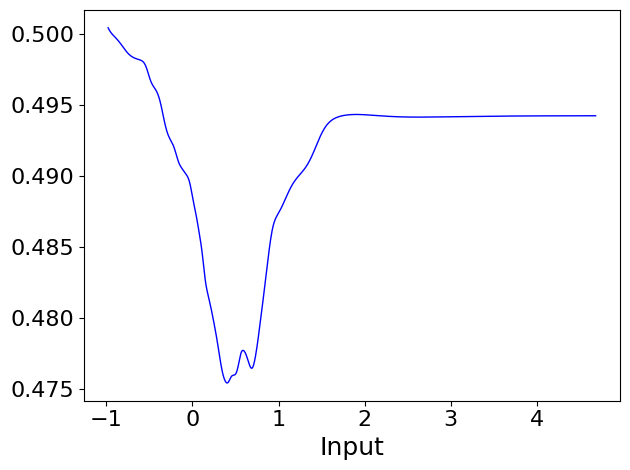}
        }
        \caption{Partial Dependence Plot - feature B} \label{fig:4}
    \end{subfigure}

    \vspace{0.5cm}

    \begin{subfigure}[b]{0.45\textwidth}
    \scalebox{1.05}{
        \includegraphics[width=\linewidth]{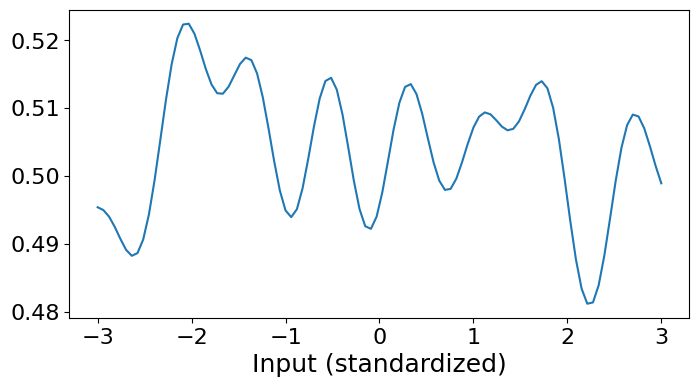}
        }
        \caption{Attribution of feature A toward the
output prediction using the Fourier KAN} \label{fig:2}
    \end{subfigure}
    \hfill
    \begin{subfigure}[b]{0.45\textwidth}
    \scalebox{1.05}{
        \includegraphics[width=\linewidth]{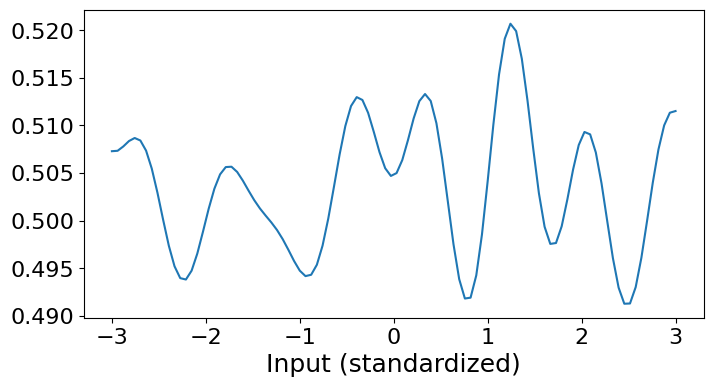}
        }
        \caption{Attribution of feature B toward the
output prediction using the Fourier KAN} \label{fig:5}
    \end{subfigure}
    \vspace{0.5cm}
\end{figure}
\begin{figure}[htbp]
    \centering
    \ContinuedFloat
    \begin{subfigure}[b]{0.45\textwidth}
        \includegraphics[width=\linewidth]{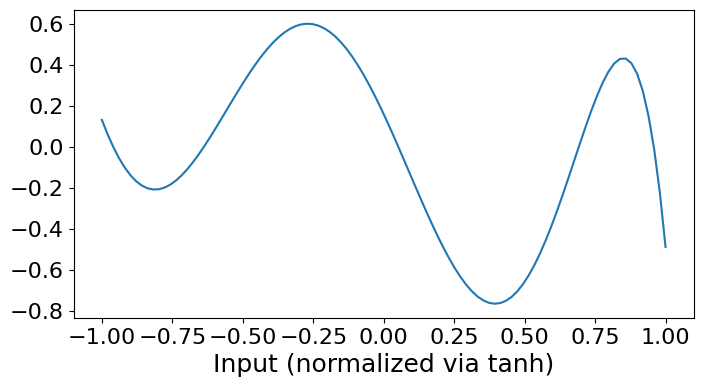}
        \caption{Attribution of feature A toward the
output prediction using the ChebyKAN} \label{fig:3}
    \end{subfigure}
    
    \begin{subfigure}[b]{0.45\textwidth}
        \centering
        \includegraphics[width=\linewidth]{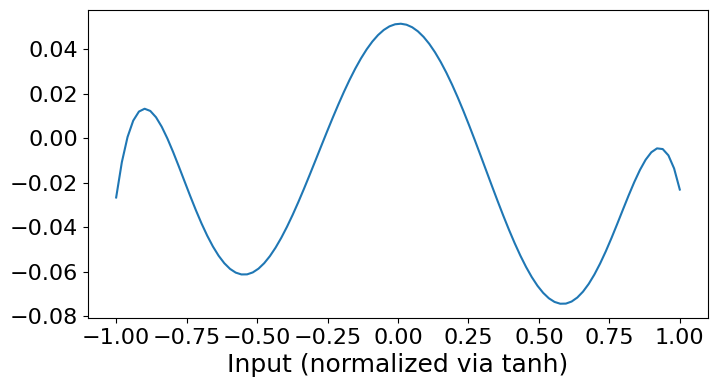}
        \caption{Attribution of feature B toward the
output prediction using the ChebyKAN} \label{fig:6}
    \end{subfigure}

  \caption{Comparison of model interpretability between the built-in function-based explanations from TabKAN and a baseline using Partial Dependence Plots (PDP). TabKAN provides direct, parameterized feature-level insights, while PDP relies on post hoc approximations that may overlook complex interactions.} 
\end{figure} 

Finally, the parametric nature of KANs ensures reproducibility in interpretation. Unlike post hoc methods (e.g., SHAP or LIME), which can vary with input perturbations, KANs provide consistent functional mappings tied directly to the model’s architecture. 


\subsection{Feature Importance and Dimensionality Reduction}
\begin{figure}
    \centering
    \includegraphics[width=1\linewidth]{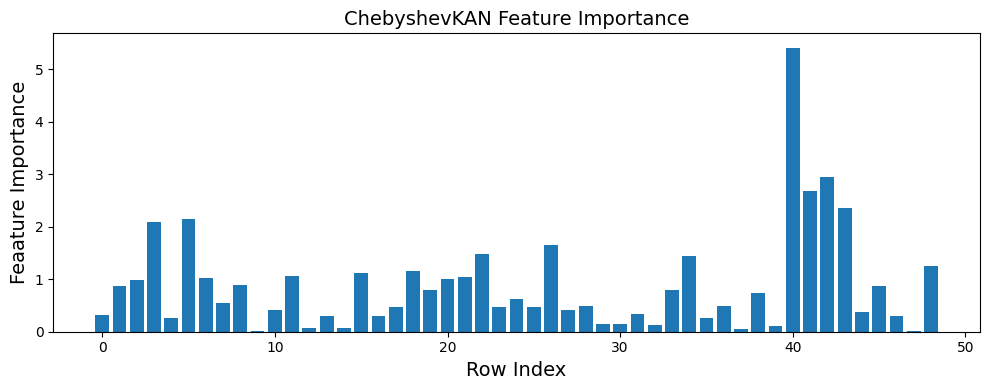}
    \caption{Feature Importance based on ChebyKAN}
    \label{fig:chebyImportance}
\end{figure}

\begin{figure}
    \centering
    \includegraphics[width=1\linewidth]{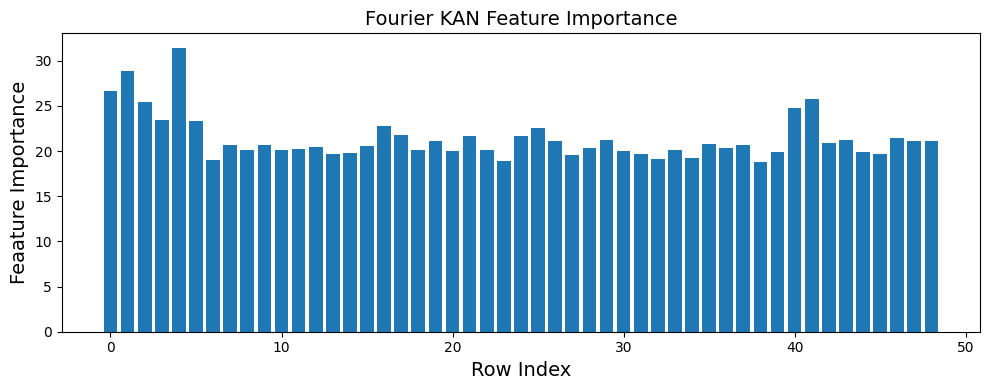}
    \caption{Feature Importance based on  Fourier KAN}
    \label{fig:forierImportance}
\end{figure}

We evaluate the feature importance and dimensionality reduction capabilities of the proposed TabKAN framework by analyzing the magnitude of coefficients derived from the Chebyshev and Fourier-based KAN equations. Specifically, we compute the absolute values of the coefficients from the Chebyshev equation in \ref{eq:cheby} and the Fourier equation in \ref{eq:fourier}. Figure \ref{fig:chebyImportance} depicts the ranked feature importance derived from the Chebyshev coefficients, while Figure \ref{fig:forierImportance} illustrates the corresponding rankings from the Fourier coefficients.

\begin{figure}[htbp]
    \centering
    \begin{minipage}{0.48\textwidth}
        \centering
        \includegraphics[width=\textwidth, height=5cm]{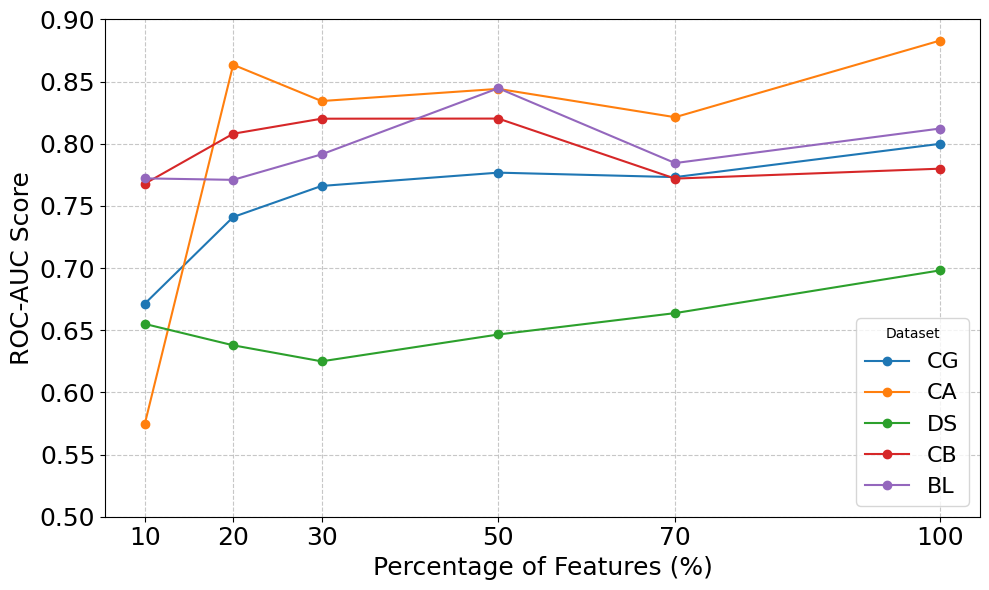}
        \caption{AUC vs Percentage of Top Selected Features for  Fourier KAN}
        \label{figs:TopFeatures_F}
    \end{minipage}
    \hfill
    \begin{minipage}{0.48\textwidth}
        \centering
        \includegraphics[width=\textwidth, height=5cm]{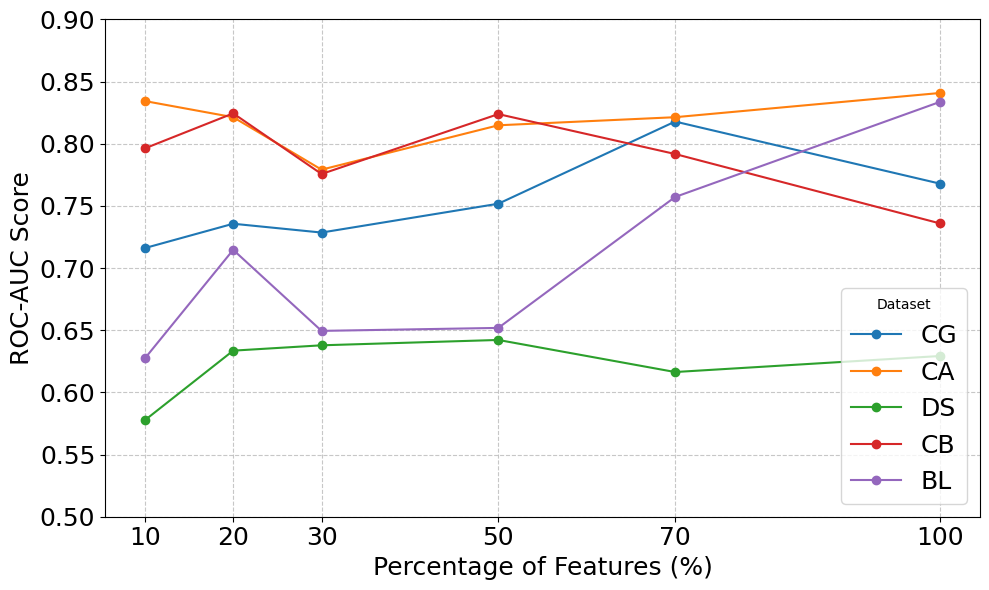}
        \caption{AUC vs Percentage of Top Selected Features for ChebyKAN}
        \label{fig:TopFeatures_C}
    \end{minipage}
\end{figure}

Based on these rankings, we conducted further experiments to assess the predictive performance of Fourier KAN and Chebyshev KAN models using subsets of features identified by their coefficients. Figures \ref{figs:TopFeatures_F} and \ref{fig:TopFeatures_C} illustrate the ROC-AUC performance across five datasets (CG, CA, DS, CB, BL) after varying levels of feature reduction. The results indicate that utilizing all available features does not necessarily yield the best predictive performance. In fact, for some datasets, models trained on reduced feature sets achieve comparable or even superior accuracy. 

\begin{figure}[ht]
    \centering
    \begin{subfigure}{\linewidth}
        \centering
        \includegraphics[width=\linewidth]{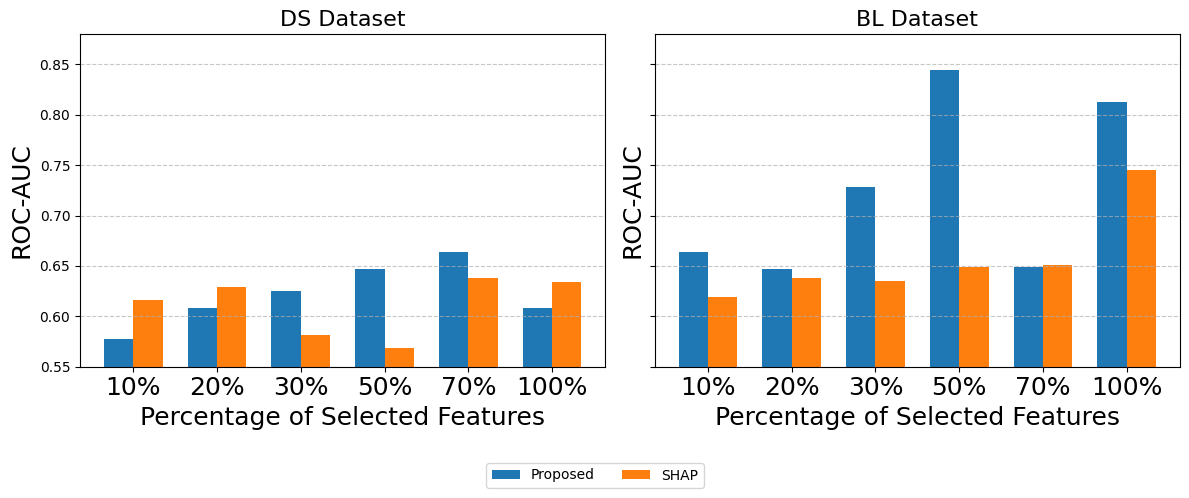}
        \caption{Comparison of Selective top important features in FourierKAN}
        \label{fig:SHAP_F}
    \end{subfigure}

    \vspace{0.5em} 

    \begin{subfigure}{\linewidth}
        \centering
        \includegraphics[width=\linewidth]{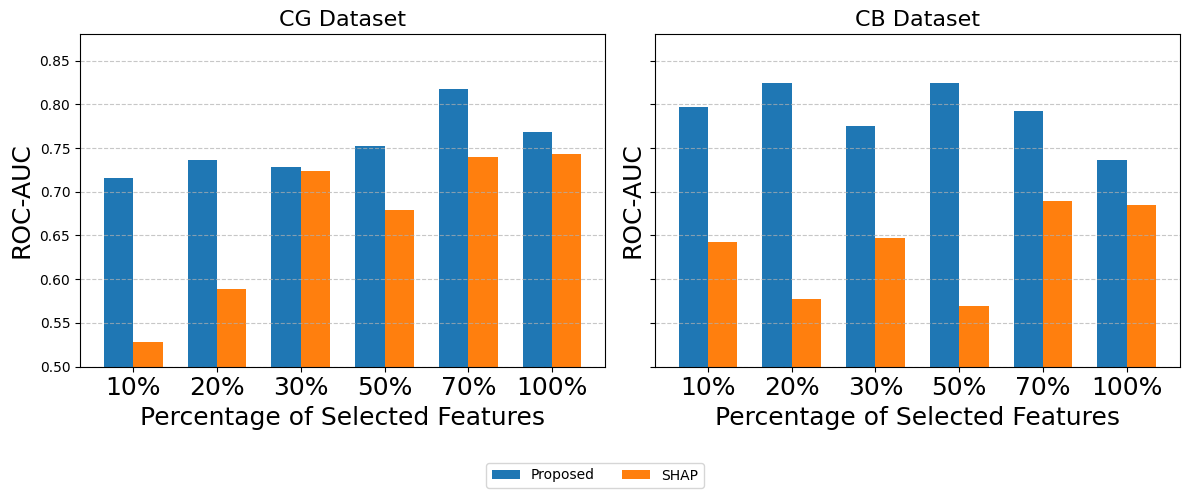}
        \caption{Comparison of Selective top important features in ChebyKAN}
        \label{fig:SHAP_C}
    \end{subfigure}

    \caption{Comparison of feature importance selection between the proposed method and SHAP across two datasets.}
    \label{fig:SHAP_combined}
\end{figure}

Figure~\ref{fig:SHAP_F} reports the AUC values obtained using various subsets of top-ranked features identified by the proposed FourierKAN-based method, compared with those selected by SHAP analysis. The results demonstrate that model-specific feature importance consistently yields superior AUC performance when less significant features are removed. Similarly, experiments conducted with ChebyKAN using the CG and CB datasets (Figure~\ref{fig:SHAP_C}) reinforce the observation that the proposed approach outperforms SHAP-based feature selection in achieving stable and improved predictive accuracy. While there is some overlap in the selected features between the SHAP-based and model-specific methods, the proposed approach often provides more stable or higher predictive performance. This outcome highlights the advantage of using learned functional parameters as a built-in mechanism for feature selection, which is both efficient and closely aligned with the model's internal representation.

\section{Ablation Study}

\subsection{Fine-tunning}
In transfer learning scenarios, where a pre-trained model is adapted to a new task or domain, the GRPO \cite{shao2024deepseekmath} framework provides a robust mechanism for fine-tuning by balancing task-specific adaptation and knowledge retention. Using a policy gradient method, GRPO optimizes model parameters $\theta$ through advantage-weighted updates derived from reward signals ($R \in \{0,1\}$), which measure the alignment between sampled predictions ($o \sim \pi_\theta$) and ground-truth labels. To address catastrophic forgetting-a typical issue in transfer learning-the method includes a Kullback-Leibler (KL) divergence penalty $\beta \cdot \mathbb{D}_{\text{KL}}(\pi_\theta \| \pi_{\text{ref}})$, which constrains deviations from the reference policy $\pi_{\text{ref}}$ (e.g., the original pre-trained model). By sampling $G$ candidate predictions per input and calculating normalized advantages $\hat{A} = R - \mathbb{E}[R]$, GRPO promotes exploration while maintaining stability, which makes it well-suited for tasks with limited target-domain data.

\begin{align}
\mathcal{J}_{\text{GRPO}}(\theta) &= \underbrace{\mathbb{E}_{\substack{q \sim \text{Batch} \\ o \sim \pi_\theta}} \left[ \frac{1}{G} \sum_{i=1}^G \log \pi_\theta(o_i|q) \cdot \hat{A}_i \right]}_{\text{Policy Gradient Loss}} 
+ \underbrace{\beta \cdot \mathbb{E}_{q \sim \text{Batch}} \left[ \mathbb{D}_{\text{KL}} \left( \pi_\theta(\cdot|q) \big\| \pi_{\text{ref}}(\cdot|q) \right) \right]}_{\text{KL Divergence Penalty}} \\
\hat{A}_i &= R_i - \mathbb{E}[R_i] \quad \text{(Advantage)} \\
R_i &= \begin{cases} 
1 & \text{if prediction } o_i = \text{label} \\
0 & \text{otherwise}
\end{cases} \\
\mathbb{D}_{\text{KL}}(\pi_\theta \| \pi_{\text{ref}}) &= \sum_{c \in \{0,1\}} \pi_\theta(c|q) \log \frac{\pi_\theta(c|q)}{\pi_{\text{ref}}(c|q)}
\end{align}

\begin{align}
    \mathcal{J}_{\text{GRPO}}(\theta) = -\mathbb{E}\left[\log \pi_\theta(o|q) \cdot \hat{A}\right] + \beta \cdot \mathbb{E}\left[\mathbb{D}_{\text{KL}}(\pi_\theta \| \pi_{\text{ref}})\right]
\end{align}

\subsection{Ablation on Enhanced Architecture}
We conducted an ablation study to evaluate the effectiveness of the KAN-Mixer architecture. As shown in Table~\ref{table:mixer}, several KAN-Mixer variants, including ChebyKAN-Mixer, JacobiKAN-Mixer, and FourierKAN-Mixer, demonstrate improved performance over both the standard KAN-based models and the original MLP-Mixer across specific datasets. 
The MLP-Mixer results used for comparison were obtained from Table 2 of~\cite{eslamian2025tabmixer}. The ablation study confirms the potential of hybrid designs that embed functional approximators like KAN within structured deep learning architectures.

\begin{table*}[htbp]
\centering
\caption{Evaluation of Different Enhanced Models for Supervised Learning} \label{table:mixer}
\begin{tabular}{l|cccccc}
\hline
Methods & CG & CA & DS & CB & BL & IO \\
\hline

\textbf{ChebyKAN-Mixer} & 0.824$^*$ & 0.863 & 0.706$^*$ & 0.807 & 0.832 & 0.950  \\

\textbf{JacobiKAN-Mixer} & 0.817 & 0.876$^*$ & 0.715$^*$ & 0.767 & 0.843$^*$  & 0.950 \\

\textbf{Fourier KAN-Mixer} & 0.850$^*$ & 0.909$^*$ & 0.715$^*$  & 0.826$^*$ & 0.707$^*$ & 0.914 \\

\midrule
\end{tabular}
\end{table*}

\subsection{Ablation on Feature Scaling and Distribution}

We conducted an ablation on input scaling and marginal distributions across three datasets (CG, IO, AD) and four TabKAN variants (ChebyKAN, fastKAN, FourierKAN, fKAN). Three preprocessing modes were compared using identical splits and hyperparameters: \emph{raw} (no scaling), \emph{standardized} (z-score), and \emph{quantile} (rank Gaussian). Overall, TabKAN variants are robust to feature scale/distribution, with standardized or quantile preprocessing offering small but consistent gains on CG and IO, and negligible changes on AD. For example, on CG, ChebyKAN test AUC improves from \(\,0.794\to0.854\to0.882\,\) (raw\(\to\)standard\(\to\)quantile), and test Acc from \(0.761\to0.779\to0.811\). On IO, ChebyKAN rises from AUC \(0.954\) (raw) to \(0.972\) (standard) with a parallel Acc gain \(0.923\to0.942\); fastKAN/FourierKAN show similar trends. On AD, all ChebyKAN settings are within \(\approx 0.01\) AUC and \(\approx 0.01\) Acc, indicating limited sensitivity at larger scale. We also observed occasional instability without scaling (e.g., fKAN on AD in raw mode producing NaNs), which disappears under standardization. In practice, we recommend standardized inputs as a default, with quantile transforms yielding additional improvements on smaller or more skewed datasets.

\begin{table*}[htbp]
\centering
\caption{CG dataset: validation and test performance across preprocessing modes.}
\label{tab:cg_scale_sensitivity}
\resizebox{0.8\textwidth}{!}{
\begin{tabular}{llcccc}
\toprule
\textbf{Mode} & \textbf{Model} & \textbf{Val Acc} & \textbf{Val AUC} & \textbf{Test Acc} & \textbf{Test AUC} \\
\midrule
\multirow{4}{*}{Raw}
  & ChebyKAN   & 0.795    & 0.824    & 0.761    & 0.794 \\
  & fastKAN    & 0.7054   & 0.7258   & 0.7036   & 0.7749   \\
  & FourierKAN & 0.7232   & 0.8409   & 0.7429   & 0.8058   \\
  & fKAN       & 0.5000   & Nan      & 0.5000   & Nan   \\
\midrule
\multirow{4}{*}{Standard}
  & ChebyKAN   & 0.857 & 0.912 & 0.779 & 0.854 \\
  & fastKAN    & 0.8393   & 0.8965   & 0.8286   & 0.9068   \\
  & FourierKAN & 0.7857   & 0.8804   & 0.8036   & 0.8840  \\
  & fKAN       & 0.8304   & 0.8870   & 0.8036   & 0.8758   \\
\midrule
\multirow{4}{*}{Quantile}
  & ChebyKAN   & 0.839 & 0.865 & 0.811 & 0.882 \\
  & fastKAN    & 0.8214   & 0.8702  & 0.8321   & 0.8765   \\
  & FourierKAN & 0.8036   & 0.9633   & 0.9269   & 0.9614   \\
  & fKAN       & 0.8304   & 0.8740   & 0.7857   & 0.8573   \\
\bottomrule
\end{tabular}}
\end{table*}

\begin{table*}[htbp]
\centering
\caption{IO dataset: validation and test performance across preprocessing modes.}
\label{tab:io_scale_sensitivity}
\resizebox{0.8\textwidth}{!}{
\begin{tabular}{llcccc}
\toprule
\textbf{Mode} & \textbf{Model} & \textbf{Val Acc} & \textbf{Val AUC} & \textbf{Test Acc} & \textbf{Test AUC} \\
\midrule
\multirow{4}{*}{Raw}
  & ChebyKAN   & 0.934 & 0.962 & 0.923 & 0.954 \\
  & fastKAN    & 0.9336   & 0.9687  & 0.9349   & 0.9691   \\
  & FourierKAN & 0.9502   & 0.9855   & 0.9349   & 0.9660   \\
  & fKAN       & 0.9419   & 0.9677   & 0.9249   & 0.9594   \\
\midrule
\multirow{4}{*}{Standard}
  & ChebyKAN   & 0.962 & 0.981 & 0.942 & 0.972 \\
  & fastKAN    & 0.9601   & 0.9811   & 0.9449   & 0.9759   \\
  & FourierKAN & 0.9435   & 0.9643   & 0.9429   & 0.9707   \\
  & fKAN       & 0.9551   & 0.9744   & 0.9382   & 0.9706   \\
\midrule
\multirow{4}{*}{Quantile}
  & ChebyKAN   & 0.959 & 0.979 & 0.940 & 0.970 \\
  & fastKAN    & 0.9502   & 0.9832   & 0.9475   & 0.9804   \\
  & FourierKAN & 0.9286   & 0.9633   & 0.9269   & 0.9614   \\
  & fKAN       & 0.9286   & 0.9607   & 0.9223   & 0.9476   \\
\bottomrule
\end{tabular}}
\end{table*}


\begin{table*}[htbp]
\centering
\caption{AD dataset: validation and test performance across preprocessing modes.}
\label{tab:ad_scale_sensitivity}
\resizebox{0.8\textwidth}{!}{
\begin{tabular}{llcccc}
\toprule
\textbf{Mode} & \textbf{Model} & \textbf{Val Acc} & \textbf{Val AUC} & \textbf{Test Acc} & \textbf{Test AUC} \\
\midrule
\multirow{4}{*}{Raw}
  & ChebyKAN   & 0.899 & 0.968 & 0.896 & 0.966 \\
  & fastKAN    & 0.6131   & 0.6402   & 0.6091   & 0.6398   \\
  & FourierKAN & 0.9105   & 0.9739   & 0.9055   & 0.9711  \\
  & fKAN       & 0.5001   &Nan   & 0.5000   & Nan   \\
\midrule
\multirow{4}{*}{Standard}
  & ChebyKAN   & 0.909 & 0.975 & 0.909 & 0.974 \\
  & fastKAN    & 0.9004   & 0.9654   & 0.8998   & 0.9657   \\
  & FourierKAN & 0.9144   & 0.9761   & 0.9119   & 0.9750   \\
  & fKAN       & 0.8947   & 0.9583   & 0.8889   & 0.9573  \\
\midrule
\multirow{4}{*}{Quantile}
  & ChebyKAN   & 0.909 & 0.975 & 0.909 & 0.974 \\
  & fastKAN    & 0.8907   & 0.9621   & 0.8868   & 0.9585   \\
  & FourierKAN & 0.9140   & 0.9753  & 0.9110   & 0.9745   \\
  & fKAN       & 0.9001   & 0.9676   & 0.8959   & 0.9663   \\
\bottomrule
\end{tabular}}
\end{table*}

\subsection{Ablation on Interpretability-Performance Trade-off}
We vary a frequency-weighted $\ell_{2}$ penalty $\lambda$ on Chebyshev edge coefficients and evaluate two outcomes: (i) predictive performance, measured by test accuracy and AUC, and (ii) an interpretability proxy, given by the fraction of coefficient mass in higher orders (referred to as “high-order energy,” orders $\geq 3$). As $\lambda$ increases, high-order energy is strongly reduced, producing much smoother and less oscillatory univariate edge functions, while generalization remains unchanged or slightly improves. In practice, high-order energy decreases by two to four orders of magnitude (CG: $0.599 \to 2\times10^{-4}$; IO: $0.477 \to 1.4\times10^{-3}$; AD: $0.785 \to 2.6\times10^{-3}$), yet test AUC is preserved or higher (CG: $0.858 \to 0.891$ to $0.897$; IO: $0.969 \to 0.981$ to $0.982$; AD: about $0.974$ throughout), with accuracy shifts within two percentage points. It is seen that stronger smoothness regularization produces simpler and more interpretable edge functions at essentially no cost to performance. The effect is most visible for CG, moderate for IO, and negligible for the larger AD dataset.

\begin{table*}[ht]
\centering
\caption{ChebyKAN: effect of smoothness penalty $\lambda$ on test performance and high–order energy (fraction of coefficient mass in orders $\ge 3$).}
\label{tab:chebykan_lambda_sweep}
\resizebox{\textwidth}{!}{
\begin{tabular}{lcccccc}
\toprule
\multirow{2}{*}{$\lambda$} & \multicolumn{2}{c}{CG} & \multicolumn{2}{c}{IO} & \multicolumn{2}{c}{AD} \\
\cmidrule(lr){2-3}\cmidrule(lr){4-5}\cmidrule(lr){6-7}
 & Acc / AUC & High-order & Acc / AUC & High-order & Acc / AUC & High-order \\
\midrule
$0$        & 0.796 / 0.858 & 0.5991 & 0.936 / 0.969 & 0.4769 & 0.909 / 0.974 & 0.7852 \\
$10^{-6}$  & 0.807 / 0.891 & 0.0003 & 0.937 / 0.981 & 0.0165 & 0.909 / 0.974 & 0.0704 \\
$10^{-5}$  & 0.818 / 0.892 & 0.0002 & 0.940 / 0.982 & 0.0033 & 0.908 / 0.973 & 0.0132 \\
$10^{-4}$  & 0.796 / 0.897 & 0.0002 & 0.942 / 0.981 & 0.0014 & 0.908 / 0.973 & 0.0026 \\
\bottomrule
\end{tabular}}
\end{table*}

\begin{table*}[ht]
\centering
\caption{FourierKAN: effect of smoothness penalty $\lambda$ on test performance and high–frequency energy.}
\label{tab:fourierkan_lambda_sweep}
\resizebox{\textwidth}{!}{
\begin{tabular}{lcccccc}
\toprule
\multirow{2}{*}{$\lambda$} & \multicolumn{2}{c}{CG} & \multicolumn{2}{c}{IO} & \multicolumn{2}{c}{AD} \\
\cmidrule(lr){2-3}\cmidrule(lr){4-5}\cmidrule(lr){6-7}
 & Acc / AUC & High-order & Acc / AUC & High-order & Acc / AUC & High-order \\
\midrule
$0$        & 0.779 / 0.850 & 0.6208 & 0.939 / 0.975 & 0.6005 & 0.912 / 0.975 & 0.5610 \\
$10^{-6}$  & 0.779 / 0.850 & 0.6208 & 0.941 / 0.975 & 0.6005 & 0.912 / 0.975 & 0.5610 \\
$10^{-5}$  & 0.779 / 0.850 & 0.6208 & 0.939 / 0.975 & 0.6005 & 0.912 / 0.975 & 0.5829 \\
$10^{-4}$  & 0.779 / 0.850 & 0.6208 & 0.941 / 0.975 & 0.6005 & 0.912 / 0.975 & 0.5853 \\
\bottomrule
\end{tabular}}
\end{table*}


\section{Conclusion}
In this work, we introduced TabKAN, a novel Kolmogorov–Arnold Network (KAN)-based architecture specifically designed for tabular data analysis. By leveraging modular and mathematically interpretable KAN components, TabKAN achieves strong performance in both supervised and transfer learning tasks, significantly outperforming classical and Transformer-based models in knowledge transfer. Unlike conventional deep learning approaches that rely on post hoc interpretability methods, TabKAN enables \emph{built-in, model-specific interpretability}, allowing direct visualization and quantitative analysis of feature interactions within the network. To enhance expressiveness and adaptability, we further developed multiple specialized KAN variants, including ChebyKAN, JacobiKAN, PadeRKAN, FourierKAN, fKAN, and fast-KAN—each offering distinct strengths in function approximation and computational efficiency. We also introduced a novel fine-tuning strategy based on GRPO optimization to improve cross-domain knowledge transfer.

The originality of this work lies in three key aspects: 1) It presents the first \emph{systematic framework} that integrates diverse KAN variants optimized specifically for tabular data learning. 2) It introduces a dedicated transfer learning methodology with GRPO fine-tuning to address domain shifts in structured datasets. 3) It provides \emph{intrinsic interpretability} through function-level visualization, eliminating reliance on post hoc explanation methods.

These contributions establish TabKAN as a novel and interpretable alternative that bridges traditional machine learning and modern deep learning for structured data. Our experiments across multiple benchmark datasets highlight the robustness, efficiency, and scalability of KAN-based architectures. Future work will build on these advancements and focus on further optimizing KAN architectures and extending their applicability to self-supervised learning and domain adaptation. Furthermore, the incorporation of formal sensitivity analysis techniques \cite{liu2025explainable, liu2020stochastic, liu2023data} could provide a more global understanding of feature influences and complement our model-specific interpretability methods. Such efforts will continue to support broader adoption of KANs in real-world applications, including promising future directions like Physics-Informed Neural Networks (PINNs) where the symbolic nature of KANs is a distinct advantage \cite{liu2024multi}.

\section{Acknowledgements}
We thank the creators of the public datasets and the authors of the baseline models for making these resources available for research.
We gratefully acknowledge Brian Gold, PhD, and the Gold Lab at the University of Kentucky for their support and for providing the facilities necessary to carry out this research.
This research is supported in part by the NSF under Grant IIS 2327113 and the NIH under Grants R21AG070909, P30AG072946, and R01HD101508-01. 

\noindent \textbf{Declaration of generative AI and AI-assisted technologies in the writing process:}
During the preparation of this work the author(s) used ChatGPT from OpenAI in order to check the grammar and improve the clarity and readability of the paper. 
After using this tool/service, the author(s) reviewed and edited the content as needed and take(s) full responsibility for the content of the publication.

\bibliographystyle{unsrt}
\bibliography{references}



\begin{appendices}

\section{Hyperparameter Sensitivity} \label{appendixA}
This appendix provides a detailed analysis of the hyperparameter sensitivity across seven neural network models (fastKAN, RKAN, fKAN, ChebyKAN, KAN, FourierKAN) evaluated on eight datasets (IO, IC, DS, CG, CB, CA, BL, AD). The analysis focuses on four key architectural metrics: Layers, Neurons, Order, and Grid, as summarized in Tables~\ref{tbl:h_io}, \ref{tbl:h_ic}, \ref{tbl:h_ds}, \ref{tbl:h_cg}, \ref{tbl:h_cb}, \ref{tbl:h_ca}, \ref{tbl:h_bl}, and \ref{tbl:h_ad}.

The architectural complexity of the models varies significantly, with distinct patterns that emerge in terms of depth, width, and approximation strategies. The RKAN and fKAN models consistently employ the highest number of layers; RKAN reaches up to 7.7 layers in the BL dataset and fKAN averages 7.5 layers in the IC and CA datasets. Such a design suggests a reliance on depth to capture complex patterns. In contrast, fastKAN and ChebyKAN use fewer layers, typically ranging from 1.5 to 3.5 on average, which favors simpler architectures. The variability in the number of layers is particularly high for RKAN and ChebyKAN, as indicated by their large standard deviations (e.g., ChebyKAN: std=2.6 in DS), which reflects dataset-specific adjustments in depth.

In terms of width, ChebyKAN consistently uses the most neurons, with means that range from 114 to 134 across datasets, followed by fastKAN, which averages between 105 and 149 neurons. This fact indicates a preference for wide, high-capacity layers. On the other hand, KAN and FourierKAN are the most compact. KAN averages 11.7 to 41.1 neurons and FourierKAN averages 33.1 to 46.9 neurons. The stability of neuron counts also varies across models. KAN exhibits low variability (std=3.8–8.4), which suggests consistent architectural choices, while ChebyKAN and fastKAN show high variability (e.g., ChebyKAN: std=57.9 in BL), which indicates dataset-specific tuning.

The order of basis functions reflects the complexity of the approximation and also varies across models. ChebyKAN and fKAN use the highest-order basis functions, with ChebyKAN averaging 4.2 to 5.4 and fKAN averaging 3.0 to 4.1. The design likely supports precise approximations but may increase computational cost. In contrast, KAN uses the lowest order, averaging 1.1 to 2.9, and favors simpler models. Notably, fastKAN and FourierKAN do not use order parameters, which implies fixed or non-polynomial basis functions.

Grid-based approximations are employed by KAN and FourierKAN, and FourierKAN uses the largest grids, averaging 10.2 in the BL dataset. This fact suggests the use of grid-based methods, such as Fourier transforms or splines, to achieve adaptive resolution. The variability in grid sizes is significant, particularly for FourierKAN (std=2.6 in BL), which indicates adjustments based on dataset complexity.

Dataset-specific trends further highlight the adaptability of these models. For example, in the IO dataset, ChebyKAN uses the widest layers (mean=121 neurons), while KAN is the most efficient (mean=41.1 neurons). In the IC dataset, KAN has the smallest architecture (mean=13.7 neurons), which contrasts with fastKAN (mean=149.3 neurons). The AD dataset showcases ChebyKAN with the highest order (mean=5.4), while fastKAN has the lowest neuron count (mean=46.4) but higher depth (mean=3.5 layers). In the BL dataset, RKAN and fKAN are the deepest (mean=7.7 and 6.4 layers, respectively), while FourierKAN uses the largest grid (mean=10.2).

The trade-offs between depth, width, and approximation strategies are evident. Models like RKAN and fKAN prioritize depth, while ChebyKAN and fastKAN emphasize width. KAN strikes a balance and maintains compact architectures. The choice of approximation strategy also varies. ChebyKAN and fKAN rely on high-order polynomials for accuracy, and KAN and FourierKAN use grid-based methods. Low-variability models, such as KAN, offer consistency, while high-variability models, such as ChebyKAN, adapt to dataset complexity.

For practitioners, these insights provide guidance on model selection. ChebyKAN or fastKAN are suitable for high-dimensional data because of their wide layers and high capacity. KAN and FourierKAN are ideal for efficiency because of their compact architectures and grid-based approximations. For tasks that require the capture of complex patterns, RKAN and fKAN use depth and high-order approximations effectively.

\begin{table}[ht]
\begin{tabular}{@{}lcccccccc@{}}
\toprule
Model & \multicolumn{2}{c}{Layers} & \multicolumn{2}{c}{Neurons} & \multicolumn{2}{c}{Order} & \multicolumn{2}{c}{Grid} \\ \midrule
 & Mean & Std. & Mean & Std. & Mean & Std. & Mean & Std. \\ \cmidrule(l){2-9} 
fastKAN & 1.3 & 0.5 & 144.5 & 39.2 & - & - & - & - \\
JacobiRKAN & 1.5 & 0.8 & 77.0 & 13.0 & 2.2 & 0.4 & - & - \\
PadéRKAN & 2.8 & 1.4 & 124.7 & 30.1 & (5.0, 2.3) & (0.6, 0) & - & - \\
fKAN & 4.9 & 1.1 & 71.1 & 11.1 & 3.9 & 0.6 & - & - \\
ChebyKAN & 2.1 & 0.3 & 123.2 & 25.1 & 4.9 & 0.3 & - & - \\
KAN & 1.0 & 0.0 & 40.0 & 0.0 & 1.0 & 0.0 & 7.0 & 0.0 \\
FourierKAN & 2.6 & 0.6 & 37.2 & 6.3 & - & - & 1.9 & 0.6 \\\bottomrule
\end{tabular}
\caption{IO Dataset} \label{tbl:h_io}
\end{table}

\begin{table}[ht]
\begin{tabular}{@{}lcccccccc@{}}
\toprule
Model & \multicolumn{2}{c}{Layers} & \multicolumn{2}{c}{Neurons} & \multicolumn{2}{c}{Order} & \multicolumn{2}{c}{Grid} \\ \midrule
 & Mean & Std. & Mean & Std. & Mean & Std. & Mean & Std. \\ \cmidrule(l){2-9} 
fastKAN & 2.4 & 0.7 & 156.2 & 20.5 & - & - & - & - \\
JacobiRKAN & 1.9 & 1.5 & 19.0 & 10.0 & 3.7 & 0.6 & - & - \\
PadéRKAN & 4.0 & 1.2 & 98.0 & 39.6 & (4.5, 2.9) & (0.5, 1) & - & - \\
fKAN & 7.6 & 1.8 & 43.3 & 7.9 & 3.9 & 0.4 & - & - \\
ChebyKAN & 1.0 & 0.0 & 141.4 & 36.4 & 5.1 & 0.6 & - & - \\
KAN & 2.0 & 0.0 & 10.0 & 0.0 & 1.0 & 0.0 & 5.0 & 0.0 \\
FourierKAN & 1.0 & 0.2 & 52.9 & 10.1 & - & - & 4.5 & 3.0 \\ \bottomrule
\end{tabular}
\caption{IC Dataset}\label{tbl:h_ic}
\end{table}
\begin{table}[ht]

\begin{tabular}{@{}lcccccccc@{}}
\toprule
Model & \multicolumn{2}{c}{Layers} & \multicolumn{2}{c}{Neurons} & \multicolumn{2}{c}{Order} & \multicolumn{2}{c}{Grid} \\ \midrule
 & Mean & Std. & Mean & Std. & Mean & Std. & Mean & Std. \\ \cmidrule(l){2-9} 
fastKAN & 1.4 & 0.7 & 105.8 & 35.9 & - & - & - & - \\
JacobiRKAN & 4.6 & 2.3 & 59.1 & 11.1 & 3.2 & 0.7 & - & - \\
PadéRKAN & 10.7 & 7.0 & 92.5 & 29.3 & (3.9, 3.7) & (0.3, 1) & - & - \\
fKAN & 7.4 & 1.8 & 58.6 & 7.6 & 4.0 & 0.4 & - & - \\
ChebyKAN & 2.1 & 0.6 & 125.1 & 40.2 & 4.6 & 0.7 & - & - \\
KAN & 3.6 & 0.5 & 20.4 & 2.8 & 3.0 & 0.0 & 3.0 & 0.0 \\
FourierKAN & 2.1 & 0.4 & 39.4 & 8.1 & - & - & 6.6 & 1.5 \\\bottomrule
\end{tabular}
\caption{DS Dataset} \label{tbl:h_ds}
\end{table}
\begin{table}[ht]

\begin{tabular}{@{}lcccccccc@{}}
\toprule
Model & \multicolumn{2}{c}{Layers} & \multicolumn{2}{c}{Neurons} & \multicolumn{2}{c}{Order} & \multicolumn{2}{c}{Grid} \\ \midrule
 & Mean & Std. & Mean & Std. & Mean & Std. & Mean & Std. \\ \cmidrule(l){2-9} 
fastKAN & 1.7 & 0.8 & 116.0 & 27.6 & - & - & - & - \\
JacobiRKAN & 1.4 & 0.8 & 86.9 & 13.6 & 2.4 & 0.6 & - & - \\
PadéRKAN & 3.0 & 1.3 & 126.7 & 27.6 & (4.8, 3.2) & (0.7, 0) & - & - \\
fKAN & 3.3 & 1.9 & 70.7 & 12.9 & 2.9 & 0.5 & - & - \\
ChebyKAN & 2.6 & 0.5 & 116.0 & 20.4 & 4.3 & 0.7 & - & - \\
KAN & 3.0 & 0.0 & 26.7 & 0.0 & 3.0 & 0.0 & 5.0 & 0.0 \\
FourierKAN & 2.4 & 0.7 & 38.4 & 9.3 & - & - & 2.1 & 1.0 \\
\bottomrule
\end{tabular} 
\caption{CG Dataset} \label{tbl:h_cg}
\end{table}

\begin{table}[ht]

\begin{tabular}{@{}lcccccccc@{}}
\toprule
Model & \multicolumn{2}{c}{Layers} & \multicolumn{2}{c}{Neurons} & \multicolumn{2}{c}{Order} & \multicolumn{2}{c}{Grid} \\ \midrule
 & Mean & Std. & Mean & Std. & Mean & Std. & Mean & Std. \\ \cmidrule(l){2-9} 
fastKAN & 1.3 & 0.6 & 113.7 & 34.7 & - & - & - & - \\
JacobiRKAN & 2.8 & 2.1 & 70.2 & 16.3 & 3.0 & 0.3 & - & - \\
PadéRKAN & 15.2 & 5.7 & 105.7 & 8.9 & (4.0, 4.3) & (0.2, 0) & - & - \\
fKAN & 3.4 & 1.0 & 44.4 & 11.0 & 3.2 & 0.6 & - & - \\
ChebyKAN & 2.5 & 0.5 & 122.1 & 30.9 & 4.4 & 0.5 & - & - \\
KAN & 3.0 & 0.0 & 10.0 & 0.0 & 1.0 & 0.0 & 5.0 & 0.0 \\
FourierKAN & 2.3 & 0.5 & 34.7 & 11.9 & - & - & 2.6 & 0.8 \\
\bottomrule
\end{tabular}
\caption{CB Dataset} \label{tbl:h_cb}
\end{table}

\begin{table}[ht]
\begin{tabular}{@{}lcccccccc@{}}
\toprule
Model & \multicolumn{2}{c}{Layers} & \multicolumn{2}{c}{Neurons} & \multicolumn{2}{c}{Order} & \multicolumn{2}{c}{Grid} \\ \midrule
 & Mean & Std. & Mean & Std. & Mean & Std. & Mean & Std. \\ \cmidrule(l){2-9} 
fastKAN & 2.8 & 0.7 & 126.9 & 23.6 & - & - & - & - \\
JacobiRKAN & 1.2 & 0.8 & 64.3 & 23.5 & 2.2 & 0.6 & - & - \\
PadéRKAN & 3.4 & 2.0 & 99.9 & 24.0 & (5.0, 3.6) & (0.6, 1) & - & - \\
fKAN & 7.4 & 1.5 & 56.9 & 8.3 & 3.9 & 0.4 & - & - \\
ChebyKAN & 2.0 & 0.8 & 118.4 & 31.0 & 2.3 & 0.5 & - & - \\
KAN & 6.7 & 0.5 & 26.0 & 1.6 & 1.0 & 0.1 & 1.0 & 0.2 \\
FourierKAN & 2.3 & 0.5 & 32.4 & 7.1 & - & - & 4.8 & 1.3 \\
\bottomrule
\end{tabular}
\caption{CA Dataset} \label{tbl:h_ca}
\end{table}

\begin{table}[ht]

\begin{tabular}{@{}lcccccccc@{}}
\toprule
Model & \multicolumn{2}{c}{Layers} & \multicolumn{2}{c}{Neurons} & \multicolumn{2}{c}{Order} & \multicolumn{2}{c}{Grid} \\ \midrule
 & Mean & Std. & Mean & Std. & Mean & Std. & Mean & Std. \\ \cmidrule(l){2-9} 
fastKAN & 2.5 & 0.7 & 113.8 & 23.5 & - & - & - & - \\
JacobiRKAN & 8.0 & 1.5 & 70.3 & 7.3 & 3.5 & 0.4 & - & - \\
PadéRKAN & 2.3 & 1.6 & 137.3 & 32.2 & (4.4, 2.4) & (0.7, 1) & - & - \\
fKAN & 6.4 & 0.9 & 66.2 & 8.3 & 3.6 & 0.4 & - & - \\
ChebyKAN & 1.1 & 0.3 & 130.3 & 64.5 & 4.0 & 1.4 & - & - \\
KAN & 2.0 & 0.2 & 35.4 & 3.3 & 1.0 & 0.1 & 6.1 & 1.1 \\
FourierKAN & 1.0 & 0.1 & 35.1 & 9.7 & - & - & 11.0 & 2.3 \\
\bottomrule
\end{tabular}
\caption{BL Dataset} \label{tbl:h_bl}
\end{table}

\begin{table}[ht]

\begin{tabular}{@{}lcccccccc@{}}
\toprule
Model & \multicolumn{2}{c}{Layers} & \multicolumn{2}{c}{Neurons} & \multicolumn{2}{c}{Order} & \multicolumn{2}{c}{Grid} \\ \midrule
 & Mean & Std. & Mean & Std. & Mean & Std. & Mean & Std. \\ \cmidrule(l){2-9} 
fastKAN & 3.6 & 1.8 & 35.8 & 22.1 & - & - & - & - \\
JacobiRKAN & 3.5 & 1.2 & 24.2 & 10.9 & 2.9 & 0.4 & - & - \\
PadéRKAN & 2.7 & 1.4 & 121.1 & 26.9 & (3.8, 4.4) & (0.8, 1) & - & - \\
fKAN & 5.4 & 1.8 & 33.0 & 7.3 & 3.3 & 0.8 & - & - \\
ChebyKAN & 1.0 & 0.2 & 44.9 & 26.5 & 5.7 & 0.8 & - & - \\
KAN & 4.3 & 0.7 & 24.3 & 1.8 & 2.0 & 0.0 & 3.0 & 0.0 \\
FourierKAN & 1.0 & 0.2 & 44.9 & 8.1 & - & - & 8.8 & 2.7 \\
\bottomrule
\end{tabular}
\caption{AD Dataset} \label{tbl:h_ad}
\end{table}

\subsection{Search Sensitivity Convergence}
Figure \ref{sensivity} presents the best validation AUC obtained by TabKAN as the number of Optuna trials increases, evaluated on eight datasets. It demonstrates how model performance improves with a larger hyperparameter search budget.

\begin{figure}[htbp]
    \centering
    \begin{subfigure}[b]{0.45\textwidth} 
        \includegraphics[width=\linewidth]{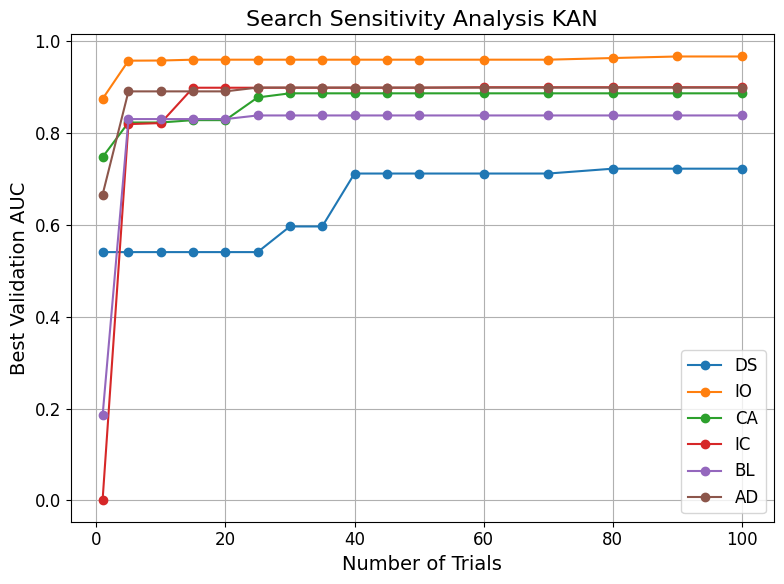}
        \caption{Search sensitivity trials for KAN convergence. Most tasks converge quickly within 15–20 trials, while DS improves gradually, indicating higher sensitivity to the search budget.}
    \end{subfigure}
    \hfill
    \begin{subfigure}[b]{0.45\textwidth}
        \includegraphics[width=\linewidth]{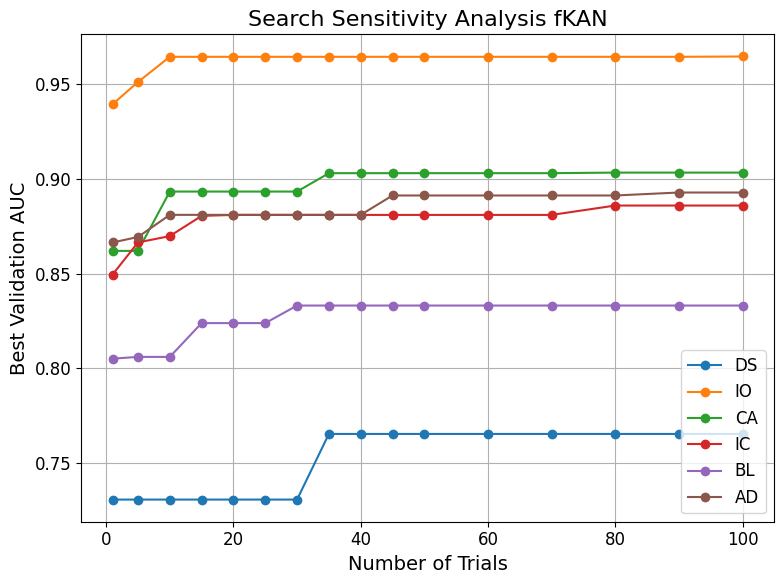}
        \caption{Search sensitivity trials for fKAN convergence. Most tasks reach stable performance within 15–30 trials. AD improves more slowly, indicating greater sensitivity to search budget.}
    \end{subfigure}

    \vspace{0.5cm}

    \begin{subfigure}[b]{0.45\textwidth}
        \includegraphics[width=\linewidth]{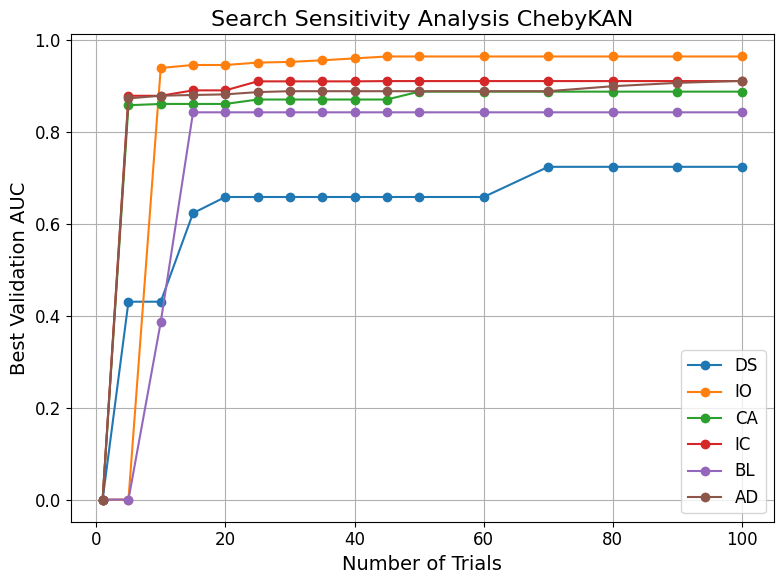}
        \caption{Search sensitivity trials for ChebyKAN convergence. Most tasks stabilize rapidly within 10–20 trials. DS requires more trials to reach optimal performance, indicating higher search sensitivity.}
    \end{subfigure}
    \hfill
    \begin{subfigure}[b]{0.45\textwidth}
        \includegraphics[width=\linewidth]{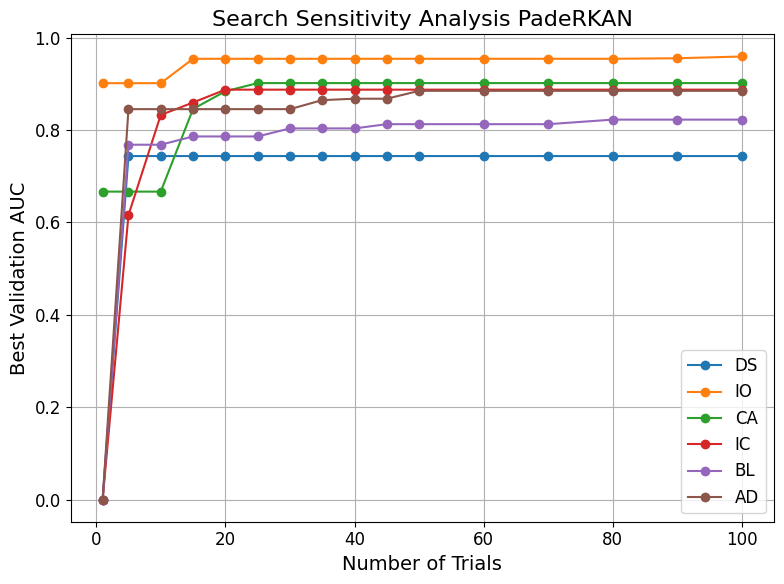}
        \caption{PadeRKAN converges rapidly, best AUC is reached within 15–20 trials, and stays consistent as more trials are run}
    \end{subfigure}
    \vspace{0.5cm}
\end{figure}

\begin{figure}[htbp]
    \centering
    \ContinuedFloat
    \begin{subfigure}[b]{0.45\textwidth}
        \includegraphics[width=\linewidth]{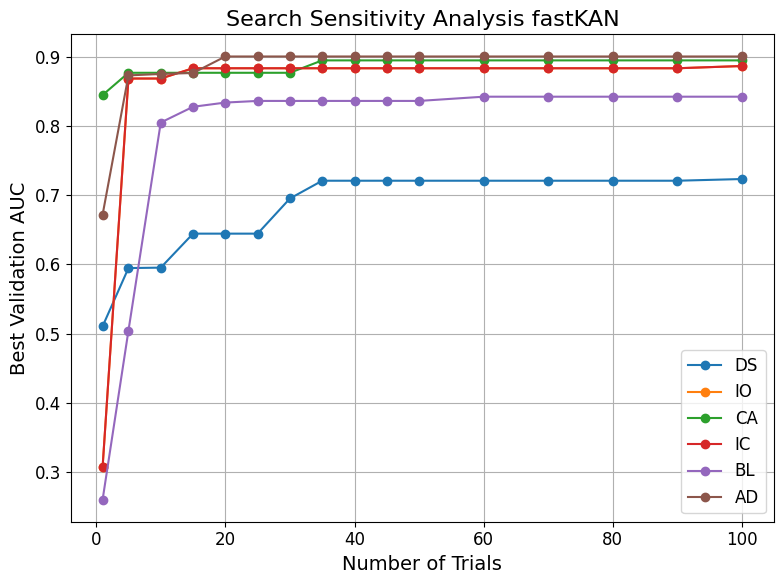}
        \caption{fastKAN converges quickly within 15–20 trials and remains consistent with additional trials. DS shows gradual improvement, indicating higher sensitivity to search budget.}
    \end{subfigure}
    \hfill
    \begin{subfigure}[b]{0.45\textwidth}
        \includegraphics[width=\linewidth]{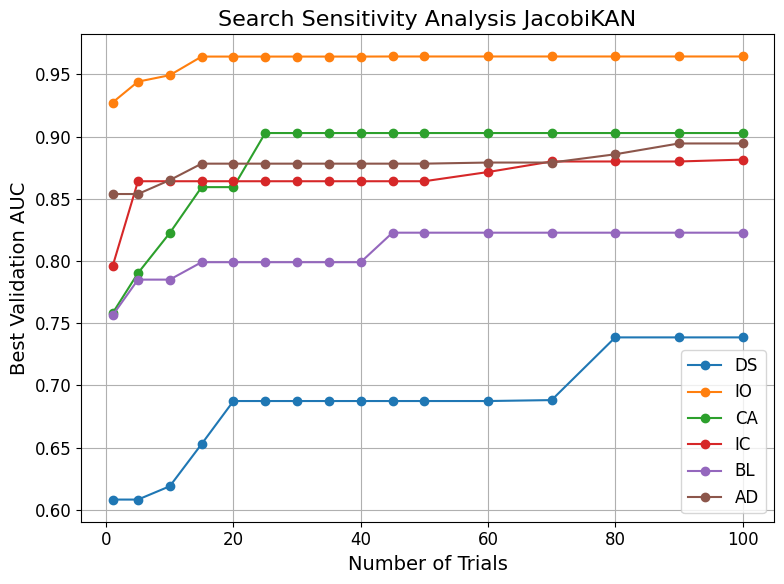}
         \caption{JacobiKAN converges rapidly within 15–20 trials and maintains stable performance with further search. DS and BL show slower improvement, reflecting higher sensitivity to search budget.}
    \end{subfigure}

    \vspace{0.5cm}

    \makebox[\textwidth][c]{%
    \begin{subfigure}[b]{0.45\textwidth}
        \centering
        \includegraphics[width=\linewidth]{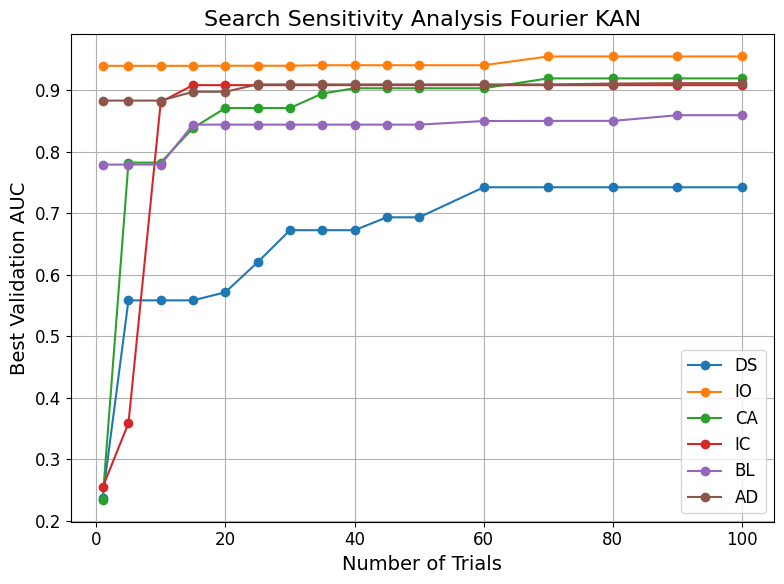}
        \caption{Fourier KAN converges within 15–20 trials and maintains stable AUC afterward. DS shows continued improvement with more trials, indicating higher sensitivity to the search budget.}
    \end{subfigure}
    }
    
    \caption{Most models converge rapidly within 15–20 trials and maintain stable validation AUC with additional trials, demonstrating search efficiency and robustness. Among all models, fastKAN, Fourier PadeRKAN, and JacobiKAN demonstrate the fastest and most stable convergence, while baseline KAN and Fourier KAN show slower improvement, particularly on the DS task, indicating greater sensitivity to the search budget.} \label{sensivity}
\end{figure}

\section{Dataset links}\label{links}
We provide the links for the public datasets that we used for the benchmark. Details of each dataset can be found in Table \ref{benchmark_datasets}. 

\begin{table}[htbp]
    \caption{Benchmark Dataset Links}
    \label{benchmark_datasets}
    \begin{tabularx}{\textwidth}{@{}lX@{}} 
        \toprule
        Dataset & URL \\
        \midrule
        Credit-G & \url{https://www.openml.org/search?type=data&status=active&id=31} \\
        Credit-Approval & \url{https://archive.ics.uci.edu/ml/datasets/credit+approval} \\
        Dress-Sales & \url{https://www.openml.org/search?type=data&status=active&id=23381} \\
        Adult & \url{https://www.openml.org/search?type=data&status=active&id=1590} \\
        Cylinder-Bands & \url{https://www.openml.org/search?type=data&status=active&id=6332} \\
        Blastchar & \url{https://www.kaggle.com/datasets/blastchar/telco-customer-churn} \\
        Insurance-Co & \url{https://archive.ics.uci.edu/ml/datasets/Insurance+Company+Benchmark+(COIL+2000)} \\
        1995-Income & \url{https://www.kaggle.com/datasets/lodetomasi1995/income-classification} \\
        ImageSegmentation  & \url{https://www.openml.org/search?type=data&sort=version&status=any&order=asc&exact_name=segment&id=36} \\
         ForestCovertype  & \url{https://archive.ics.uci.edu/dataset/31/covertype} \\ 
        \bottomrule
    \end{tabularx}
\end{table}

\section{Consistency Across 100 Seed Values}
\begin{figure}
    \centering
    \includegraphics[width=1\linewidth]{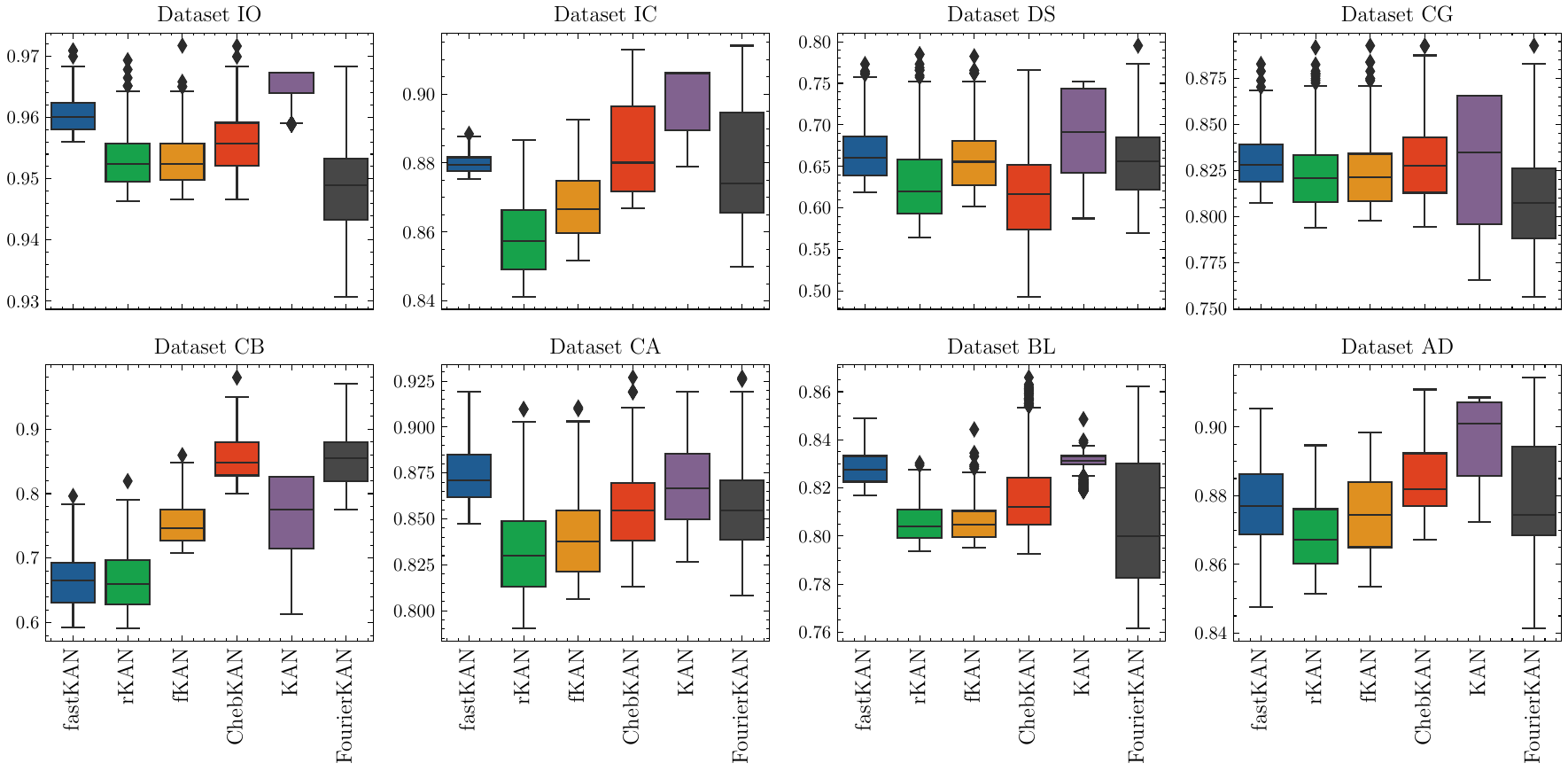}
    \caption{The interquartile range across 100 runs with varying hyperparameters highlights the influence of architecture on experimental outcomes. The plot depicts the variation in the distribution of raw and synthesized data across different training and test set splits.}
    \label{fig:enter-label}
\end{figure}

Our experiments reveal that the choice of seed plays a crucial role in influencing the results. This effect is particularly evident during the data partitioning process for training and testing. To capture this variability, we illustrate the interquartile range, offering a broader perspective on the fluctuations in our findings, as depicted in Fig. \ref{fig:enter-label}. This analysis highlights both the stability of our model and the inevitable variations in performance stemming from different data splits.

\section{List of Abbreviations}
Some abbreviations used in the main text are defined in Table \ref{abbr}.

\begin{table}[h!]
\caption{Abbreviations} \label{abbr}
\centering
\begin{tabular}{ll}
\hline
\textbf{Abbreviation} & \textbf{Full Form} \\
\hline
\multicolumn{2}{l}{\textbf{Optimization \& Algorithms}} \\
L-BFGS & Limited-memory Broyden–Fletcher–Goldfarb–Shanno \\
BFGS & Broyden–Fletcher–Goldfarb–Shanno \\
GRPO & Group Relative Policy Optimization \\
\hline
\multicolumn{2}{l}{\textbf{Machine Learning Models}} \\
KAN & Kolmogorov–Arnold Network \\
MLP & Multi-Layer Perceptron \\
SNN & Self-Normalizing Neural Network \\
DCN & Deep Cross Network \\
AutoInt & Automatic Feature Interaction via Self-Attention \\
TabNet & Attentive Interpretable Tabular Learning \\
TabTrans & TabTransformer \\
FT-Trans & Feature Tokenizer Transformer \\
VIME & Variational Information Maximizing Exploration \\
SCARF & Self-Supervised Contrastive Learning Framework for Tabular Data \\
SAINT & Self-Attention and Intersample Transformer \\
CatBoost & Categorical Boosting \\
LightGBM & Light Gradient Boosting Machine \\
XGBoost & Extreme Gradient Boosting \\
TabRet & Tabular Retokenization \\
XTab & Cross-table Pretraining for Tabular Transformers \\
TabCBM & Tabular Concept-Based Model \\
TabPFN & Tabular Prior-Data Fitted Network \\
TabMap & Tabular Topographic Map Model \\
TabSAL & Tabular Small-Agent Language Model \\
TabMixer & Tabular enhanced MLP-Mixer \\
\hline
\end{tabular}
\end{table}

\section{Preprocessing Pipeline} \label{prepocess_pipline}
To ensure complete and balanced inputs for TabKAN, we adopt the preprocessing strategy described in \cite{eslamian2025tabmixer}. Specifically, this involves a two-stage procedure: (1) imputing missing values using EM-KNN, and (2) addressing class imbalance with augmentation. The following pseudo-code provides a summarized version of the preprocessing method:

\begin{algorithm}[hb]
\caption{Tabular Data Preprocessing Pipeline}
\label{alg:preprocessing_pipeline}
\begin{algorithmic}[1]
\Require Dataset $\mathcal{D}=\{(x_i,y_i)\}_{i=1}^N$ with $x_i\in\mathbb{R}^m$, missing values, and $\min_g \left|\{\,i \mid y_i=g\,\}\right| \ll \max_g \left|\{\,i \mid y_i=g\,\}\right|$
\Ensure Balanced dataset $\mathcal{D}_{\mathrm{final}}=\{(x'_j,y'_j)\}_{j=1}^{N'}$ with no missing values

\Procedure{EM\_KNN\_Imputation}{$\mathcal{D}$}
    \For{each class $g \in \{1,\dots,G\}$}
        \State $\mathcal{D}_g \gets \{\,x_i \mid y_i=g\,\}$
        \State $X^{\mathrm{num}}_g \gets \arg\max_{\theta}\; \mathbb{E}_{z \sim p(z\mid x_{\mathrm{obs}})}\!\left[\log p(x_{\mathrm{obs}},z\mid \theta)\right]$ \Comment{EM for numerical}
        \State $X^{\mathrm{cat}}_g \gets \operatorname{mode}\!\left\{\,x_k^{\mathrm{cat}} \mid k \in \mathrm{KNN}(x_i,\mathcal{D}_g)\right\}$ \Comment{KNN for categorical}
    \EndFor
    \State \Return $\mathrm{OneHotEncode}\!\left(\bigcup_{g=1}^{G} \mathcal{D}_g\right)$
\EndProcedure

\Procedure{Balance\_Classes}{$\mathcal{D}_{\mathrm{complete}}$}
    \State $\mathcal{D}_{\mathrm{smote}} \gets \mathcal{D}_{\mathrm{complete}} \cup \left\{\,\operatorname{interpolate}(x_i,x_j) \mid x_i,x_j \in \text{minority class}\,\right\}$
    \State $\mathcal{D}_{\mathrm{vae}} \gets \mathcal{D}_{\mathrm{smote}} \cup \left\{\,x \mid z\sim\mathcal{N}(0,I),\; x \sim p(z)\right\}$ \Comment{VAE generation}
    \State $\mathcal{D}_{\mathrm{final}} \gets \mathcal{D}_{\mathrm{vae}} \cup \left\{\,x \mid x \sim q_{\phi}(x\mid z,y)\ \text{weighted by}\ \mathrm{KMM}(p_{\mathrm{data}},p_{\mathrm{model}})\right\}$ \Comment{WM-CVAE}
    \State \Return $\mathcal{D}_{\mathrm{final}}$
\EndProcedure

\State $\mathcal{D}_{\mathrm{complete}} \gets \textsc{EM\_KNN\_Imputation}(\mathcal{D})$
\State $\mathcal{D}_{\mathrm{final}} \gets \textsc{Balance\_Classes}(\mathcal{D}_{\mathrm{complete}})$
\end{algorithmic}
\end{algorithm}

\section{K-Fold Validation of TabKAN Variants} \label{k-fold}

We performed stratified $K$-fold validation ($K \in \{3,5,7\}$) on three representative datasets---CG (small), IO (medium), and AD (large)---for all three best TabKAN variants (ChebyKAN, fastKAN, fKAN) based on Table \ref{table:supervised} . Within each fold, preprocessing (imputation/encoding/scaling) was fit on the training split and applied to the validation split to prevent leakage. Algorithm \ref{alg:kfold_tabkan} represents the procedure. We used fixed hyperparameters taken from the main experiments. Table \ref{table:k-fold} report mean~$\pm$~standard deviation for Accuracy and AUROC across folds, demonstrating consistent performance of TabKAN variants across partition schemes and dataset scales.

\begin{table*}[htbp]
\centering
\caption{Comparison of different methods on CG, IO, and AD datasets with different $K$-fold settings.}
\label{table:k-fold}
\resizebox{\textwidth}{!}{
\begin{tabular}{l|ccc|ccc|ccc}
\toprule
\multirow{2}{*}{Methods} & \multicolumn{3}{c|}{CG} & \multicolumn{3}{c|}{IO} & \multicolumn{3}{c}{AD} \\
 & $k=3$ & $k=5$ & $k=7$ & $k=3$ & $k=5$ & $k=7$ & $k=3$ & $k=5$ & $k=7$ \\
\midrule
\textbf{ChebyKAN}    & $0.80_{.00}$ & $0.80_{.01}$ & $0.78_{.00}$ & $0.94_{.00}$ & $0.94_{.00}$ & $0.94_{.00}$ & $0.90_{.00}$ & $0.90_{.00}$ & $0.90_{.00}$ \\
\textbf{fastKAN}     & $0.84_{.00}$ & $0.84_{.00}$ & $0.84_{.00}$ & $0.93_{.00}$ & $0.94_{.00}$ & $0.93_{.00}$ & $0.88_{.00}$ & $0.88_{.00}$ & $0.88_{.00}$ \\
\textbf{fKAN}        & $0.81_{.00}$ & $0.82_{.01}$ & $0.79_{.01}$ & $0.94_{.00}$ & $0.94_{.00}$ & $0.94_{.00}$ & $0.87_{.00}$ & $0.87_{.00}$ & $0.87_{.00}$ \\

\bottomrule
\end{tabular}
}
\end{table*}
The results in Table~\ref{table:k-fold} show that, with the random seed fixed across all $K$ values, TabKAN variants maintain consistent accuracy across 3-, 5-, and 7-fold validation, indicating robustness of the models under different partition schemes.

\begin{algorithm}[H]
\caption{K-fold Validation for TabKAN Variants}
\label{alg:kfold_tabkan}
\begin{algorithmic}[1]
\Require Datasets $\mathcal{D} \in \{\text{CG}, \text{IO}, \text{AD}\}$, Models $\mathcal{M}=$ \{fastKAN, JacobiRKAN, PadéRKAN, fKAN, ChebyKAN, FourierKAN, KAN\}, K-folds $K \in \{3,5,7\}$
\For{each dataset $D \in \mathcal{D}$}
  \For{each $K$}
    \State Create stratified $K$-fold splits $\{(\mathcal{T}_k, \mathcal{V}_k)\}_{k=1}^{K}$
    \For{each model $M \in \mathcal{M}$}
      \For{$k = 1$ to $K$}
        \State Fit preprocessing on $\mathcal{T}_k$ (imputation/encoding/scaling)
        \State Train $M$ on preprocessed $\mathcal{T}_k$
        \State Evaluate on preprocessed $\mathcal{V}_k$; store metrics
      \EndFor
      \State Aggregate metrics: mean $\pm$ std over folds
    \EndFor
  \EndFor
\EndFor
\end{algorithmic}
\end{algorithm}

\end{appendices}
\end{document}